\documentclass[runningheads]{llncs}

\usepackage{graphicx}
\usepackage{amsmath,amssymb} 
\usepackage{color}
\usepackage[width=122mm,left=12mm,paperwidth=146mm,height=193mm,top=12mm,paperheight=217mm]
{geometry}
\usepackage{epsfig}
\usepackage[font=small,skip=2pt]{caption}
\usepackage{subfig}
\usepackage{float}
\usepackage{booktabs}
\usepackage{array,multirow}
\usepackage{arydshln}

\begin{document}
\pagestyle{headings}
\mainmatter

\title{Attentional Push: Augmenting Salience with Shared Attention Modeling} 

\titlerunning{Attentional Push: Augmenting Salience with Shared Attention Modeling}


\author{Siavash Gorji\qquad James J. Clark}
\institute{Centre for Intelligent Machines, Department of Electrical and Computer Engineering, McGill University\\
Montreal, Quebec, Canada\\
{\tt\small siagorji@cim.mcgill.ca}
{\tt\small clark@cim.mcgill.ca}}

\maketitle

\begin{abstract}
We present a novel visual attention tracking technique based on Shared Attention modeling. Our proposed method models the viewer as a participant in the activity occurring in the scene. We go beyond image salience and instead of only computing the power of an image region to pull attention to it, we also consider the strength with which other regions of the image push attention to the region in question. We use the term \textit{Attentional Push} to refer to the power of image regions to direct and manipulate the attention allocation of the viewer. An attention model is presented that incorporates the Attentional Push cues with standard image salience-based attention modeling algorithms to improve the ability to predict where viewers will fixate. Experimental evaluation validates significant improvements in predicting viewers' fixations using the proposed methodology in both static and dynamic imagery.
\end{abstract}

\section{Introduction}

Attention is a temporal selection mechanism in which a subset of available sensory information is chosen for further processing. Since the visual system cannot perform all visual functions at all locations in the visual field at the same time in parallel \cite{Tsotsos90}, attention implements a serialized mechanism that acts as an information-processing bottleneck to allow near real-time performance. Given the wider arrangement of receptors and the larger receptive fields of ganglion cells in the periphery, attention supports analysis of a scene by successively directing the high-resolution fovea to salient regions of the visual field.  
While visual attention guides the so called \textit{focus of attention} (FOA) to important parts of the scene, a key question is on the computational mechanisms underlying this guidance. 
Aside from being an interesting scientific challenge, attention tracking- determining where, and to what, people are paying attention while viewing static photographs or while watching videos and cinematic movies- has many applications in: object object detection and recognition \cite{Walther20061395}, visual surveillance \cite{Benfold2009}, human-robot interaction \cite{Ferreira2014}, and advertising \cite{Rosenholtz2011}. 

Modeling visual attention has attracted much interest recently and there are several frameworks and computational approaches available. The current state-of-the-art of attention prediction techniques are based on computing image salience maps, which provide, for each pixel, its probability to attract viewers' attention.
Almost all attention models are directly or indirectly inspired by cognitive findings. The basis of many attention models dates back to Treisman and Gelade's feature integration theory \cite{Treisman198097} which showed that during visual perception, visual features, e.g. color, size, orientation, direction of movement, brightness and spatial frequency, are registered early, automatically, and in parallel across the whole visual field. Koch and Ullman \cite{Koch85} proposed a feed-forward neural model to combine these early visual features into a central representation, i.e. the saliency map. 
Clark and Ferrier \cite{Clark98} developed a robotic vision system that used the Koch and Ullman salience model to control the motion of a binocular pair of cameras. This work was the first to demonstrate computationally the link between image salience and eye movements. Subsequently, models of salience have often been characterized by how well they predict eye movements.

Perhaps the first complete implementation of the Koch and Ullman model was proposed by the pioneering work of Itti et al. \cite{Itti98} which inspired many later models and has been the standard benchmark for comparison. This model generates feature maps across different scales for three early visual features and then linearly combines them to yield the saliency map. 
Similarly, GBVS \cite{Harel2007} extracts intensity, color, and orientation feature maps at multiple scales and builds a fully connected graph over all locations of each feature map, with weights between two nodes set proportional to the similarity of feature values and their spatial distance. The saliency map is formed by a normalized combination of the equilibrium distribution of the graphs.
Goferman et al. \cite{Goferman2012} proposed a context-aware saliency detection model. The method is based on four principles of human attention: local low-level features such as color and contrast, global considerations to maintain features that deviate from the norm, visual organization rules, and high-level factors such as human faces.
In RARE \cite{Riche2013642}, the saliency map is formed by fusing  rarity maps, which are computed using cross-scale occurrence probability of each pixel.
In AWS \cite{Garcia2012}, the local variability in energy is used as an estimation of saliency. The method decomposes the a and b color channels are into multiple scales, while decomposing the luminance channel using Gabor filter banks. The saliency map is computed as the local average of the decomposed channels.
In BMS \cite{Zhang2013}, an image is characterized by a set of binary images, generated by randomly thresholding the image's color channels. Based on a Gestalt principle of figure ground segregation, the method computes the saliency map using the topological structure of Boolean maps.

The above models only rely on bottom-up influences. While having reasonable performance, bottom-up models are mostly feed-forward, do not need training and are in general easy to apply. While many attention models fall into this category, they cannot fully explain the eye movements, since the fixations are also modulated by the visual tasks. 
In contrast to bottom-up attention, top-down attention is slow, task-driven, voluntary, uses feedback and requires learning mechanisms to be trained for a specific visual task and are therefore, more complex to deploy. Top-down attention takes higher-level cognitive cues such as task demands into account. This is probably why regardless of the important role of top-down factors in directing visual attention, the majority of existing attention models focus on bottom-up cues (see the recent extensive survey of attention modeling by Borji and Itti \cite{BorjiItti2013}). Haji-Abolhassani and Clark \cite{HajiAbolhassani2014127} developed an inverse Yarbus process in which the attention tracking system is able to infer the viewer's visual task, given the eye movement trajectories. Similar methods were proposed by Borji and Itti \cite{Borji2014} using a Boosted Classifier and by Kanan et al. \cite{Kanan2014} using a Fisher Kernel Learning method. Aside from the visual task demands, scene gist \cite{Torralba2006}, tendency of observers to look near the center of displays (also known as image center-bias \cite{Tatler2007}), and expertise with similar scenes \cite{Underwood2009}, also affect attention in a top-down manner.
 
\begin{figure}[t]   
\captionsetup[subfigure]{labelformat=empty} 
    \centering
    \null\hfill
    \subfloat{
    \includegraphics[width=0.19\textwidth]{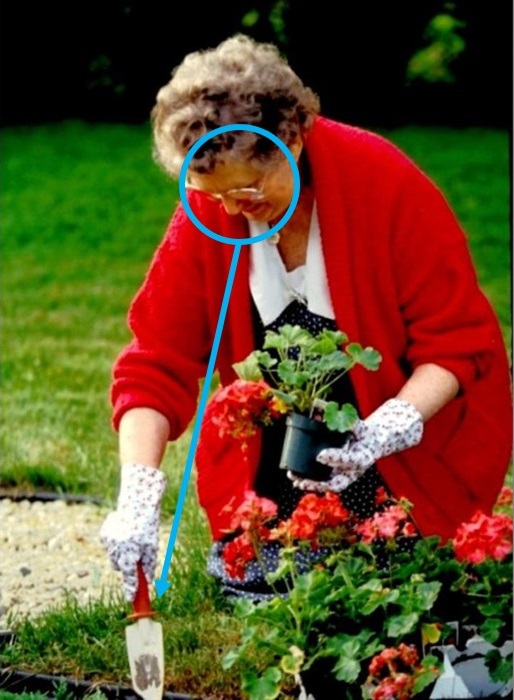}}    
    \hfill
    \subfloat{
    \includegraphics[width=0.39\textwidth]{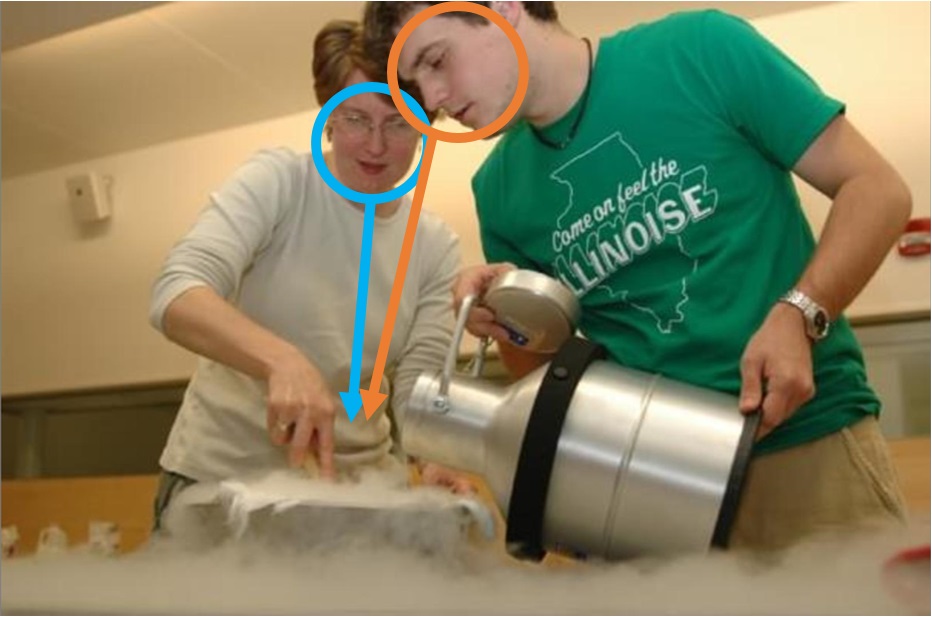}}    
    \hfill
    \subfloat{
    \includegraphics[width=0.345\textwidth]{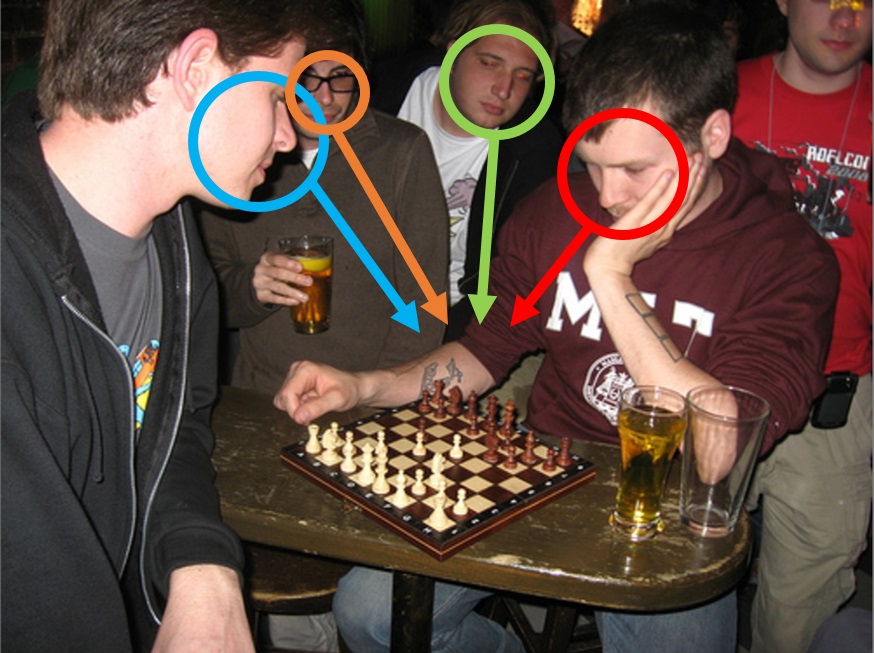}}            
	\hfill\null
\\[-1.8\baselineskip] 
	\centering
    \null\hfill
    \subfloat{
    \includegraphics[width=0.19\textwidth]{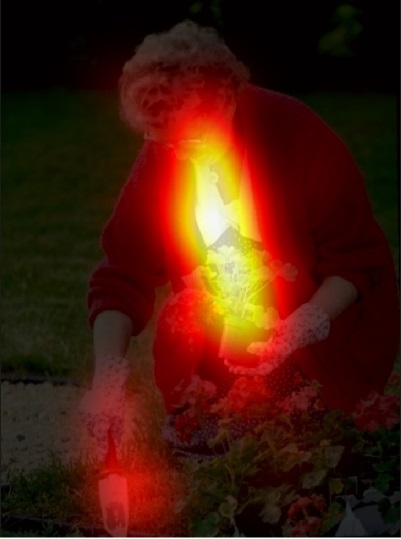}}       
    \hfill
    \subfloat{
    \includegraphics[width=0.39\textwidth]{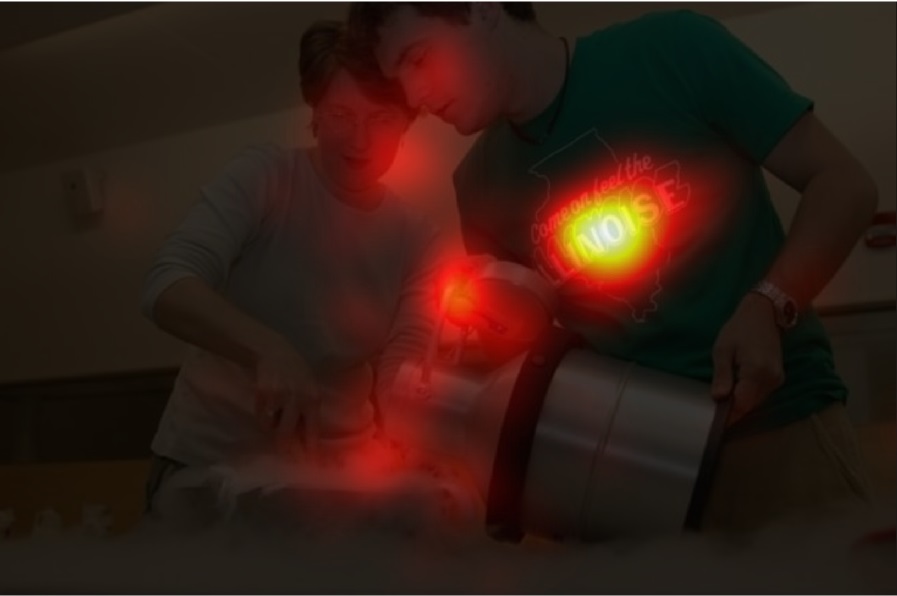}} 
    \hfill
    \subfloat{
    \includegraphics[width=0.345\textwidth]{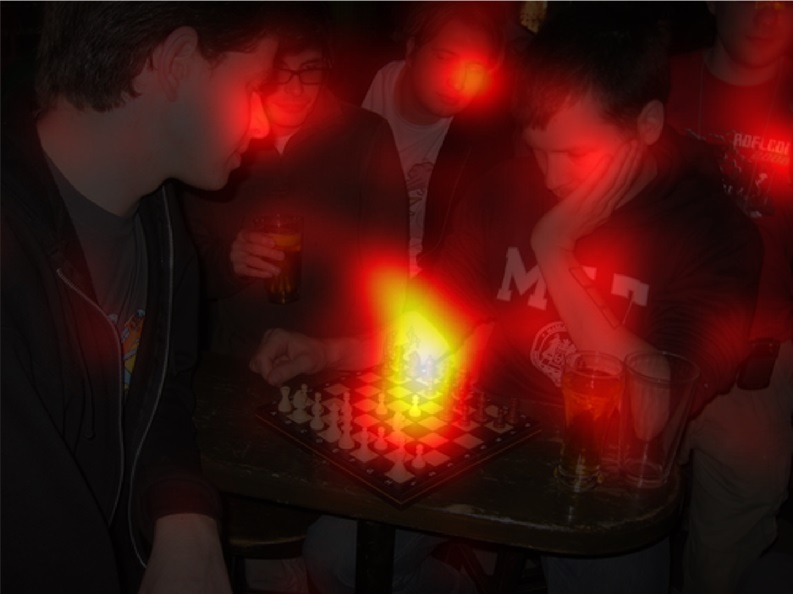}}   
	\hfill\null 
\\[-1.8\baselineskip] 
	\centering
    \null\hfill
    \subfloat{
    \includegraphics[width=0.09\textwidth]{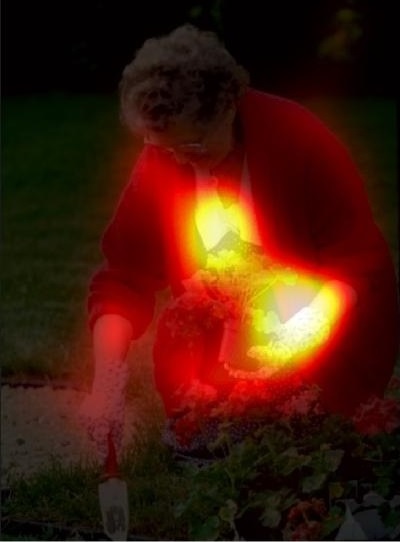}}       
    \subfloat{
    \includegraphics[width=0.09\textwidth]{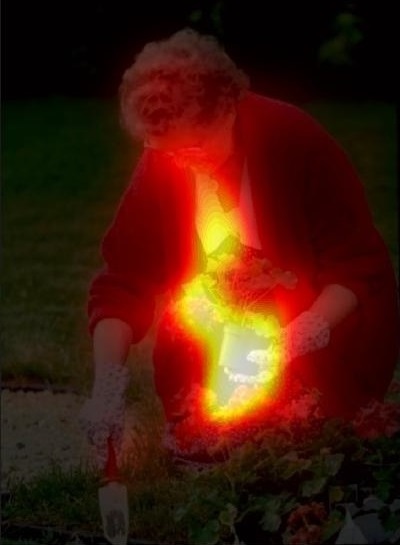}}       
    \hfill
    \subfloat{
    \includegraphics[width=0.188\textwidth]{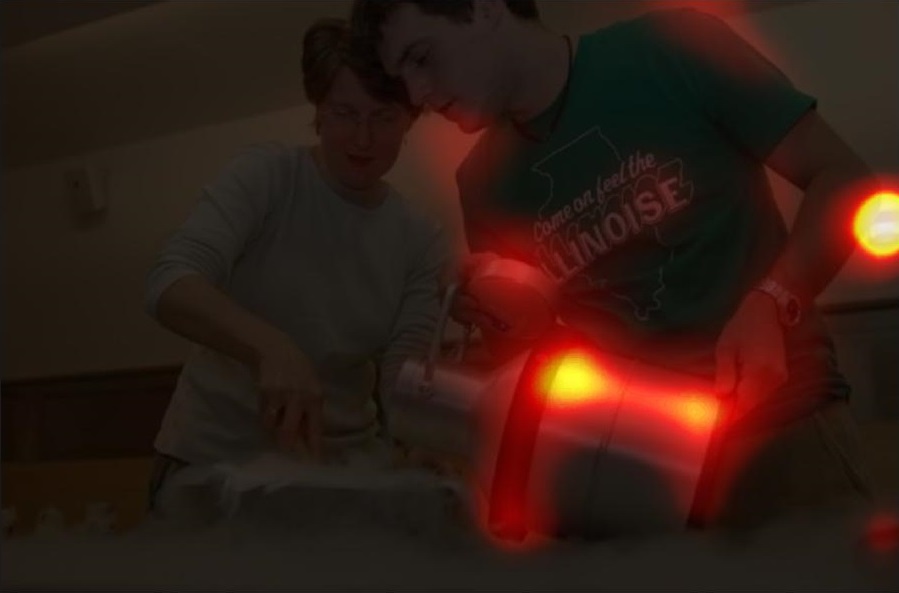}} 
    \subfloat{
    \includegraphics[width=0.19\textwidth]{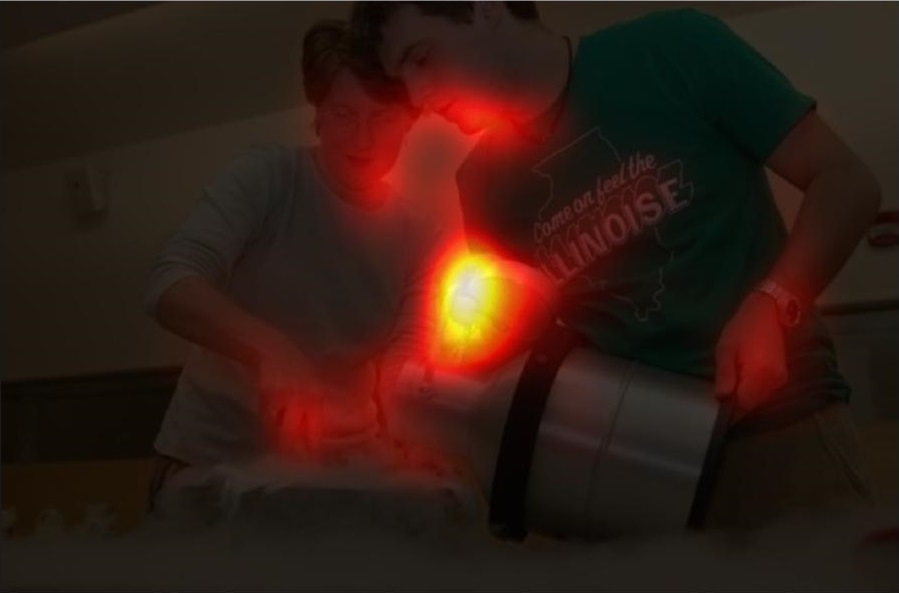}} 
    \hfill
    \subfloat{
    \includegraphics[width=0.168\textwidth]{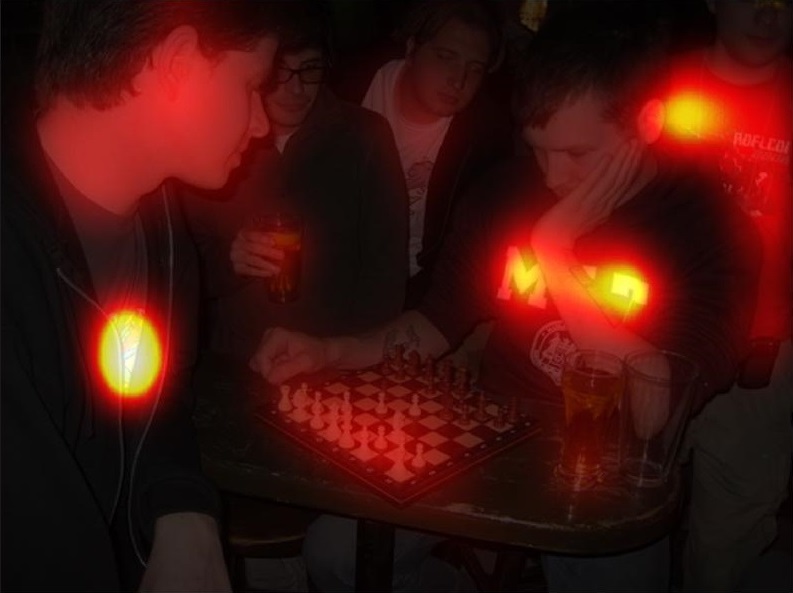}}   
    \subfloat{
    \includegraphics[width=0.168\textwidth]{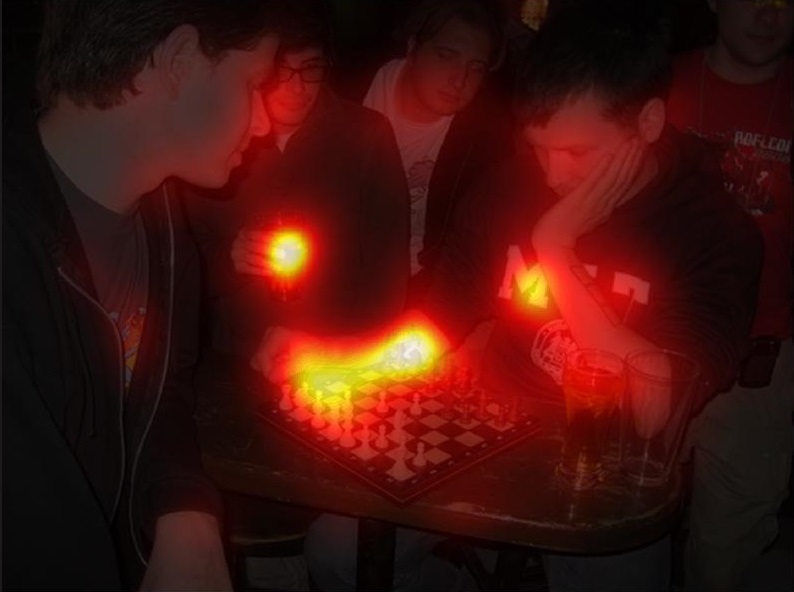}}   	
	\hfill\null 
	\setlength{\belowcaptionskip}{-20pt}
    \caption{\label{fig:Examples}
Static gaze direction as an Attentional Push cue, directing viewers' attention in social scenes. Each image has a shared locus of attention of the scene actors that has low salience, in spite of having viewers' attention allocated to them. (Top row) Original images with annotated head pose. (Middle row) Overlaid fixation maps. (Bottom row) Overlaid saliency maps: (left) BMS \cite{Zhang2015}, (right) eDN \cite{Vig2014}. The saliency maps cannot fully predict veiwers' fixations. Original images and eye fixation data are from the action and the social categories of CAT2000 dataset \cite{BorjiI15}). Saliency maps were histogram-matched to the fixation maps for visualization.}
\end{figure}

All of the aforementioned methods are based on saliency maps, and only differ in their choice of features to be used in forming the maps, and in the way top-down guidance modulates the salience. In a recent comparative study, Borji et al. \cite{Borji2013} compared 35 state-of-the-art of saliency models over synthetic and natural stimuli. They showed that these methods are far from completely predicting viewers' attentional behavior. 
A possible reason for this mediocre performance is that image salience is not the only factor driving attention allocation. 
Birmingham et al. \cite{Birmingham2009} assessed the ability of the Itti et al. \cite{Itti98} saliency map in predicting eye fixations in social scenes and showed that its performance is near chance levels. They concluded that the viewer's eye movements are affected by their interest to social information of the scenes. 
In a recent study, Borji et al. \cite{BorjiParksItti2014} investigated the effect of gaze direction on the bottom-up saliency. They conducted a controlled experiment in which an actor is asked to look at two different objects in turn, resulting in two images that differed only by the actor's gaze direction. The experiments show that the median of the fraction of all saccades that start from the head and end inside the gazed-at object to that of the ignored object is more than 3. This clearly shows that low-level saliency cannot account for the influence of gaze direction on fixations. The study also highlights that the median of the saccade directions in the actor's gaze direction is about 9.5 times higher than the chance level, which indicates that viewers tend to look more in the direction of actor's gaze than in other directions \cite{BorjiParksItti2014}.

\begin{figure}[t]   
\captionsetup[subfigure]{labelformat=empty} 
    \centering
    \null\hfill
    \subfloat{
    \includegraphics[width=0.3\textwidth]{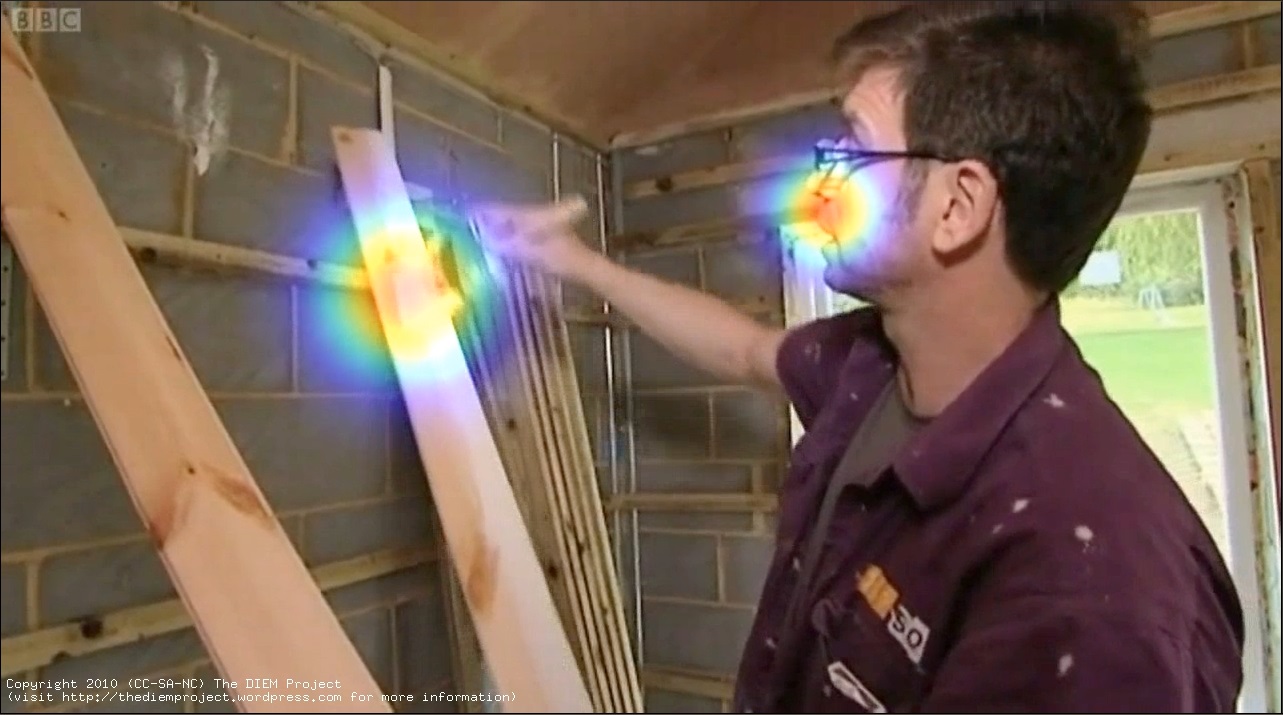}}    
    \hfill
    \subfloat{
    \includegraphics[width=0.3\textwidth]{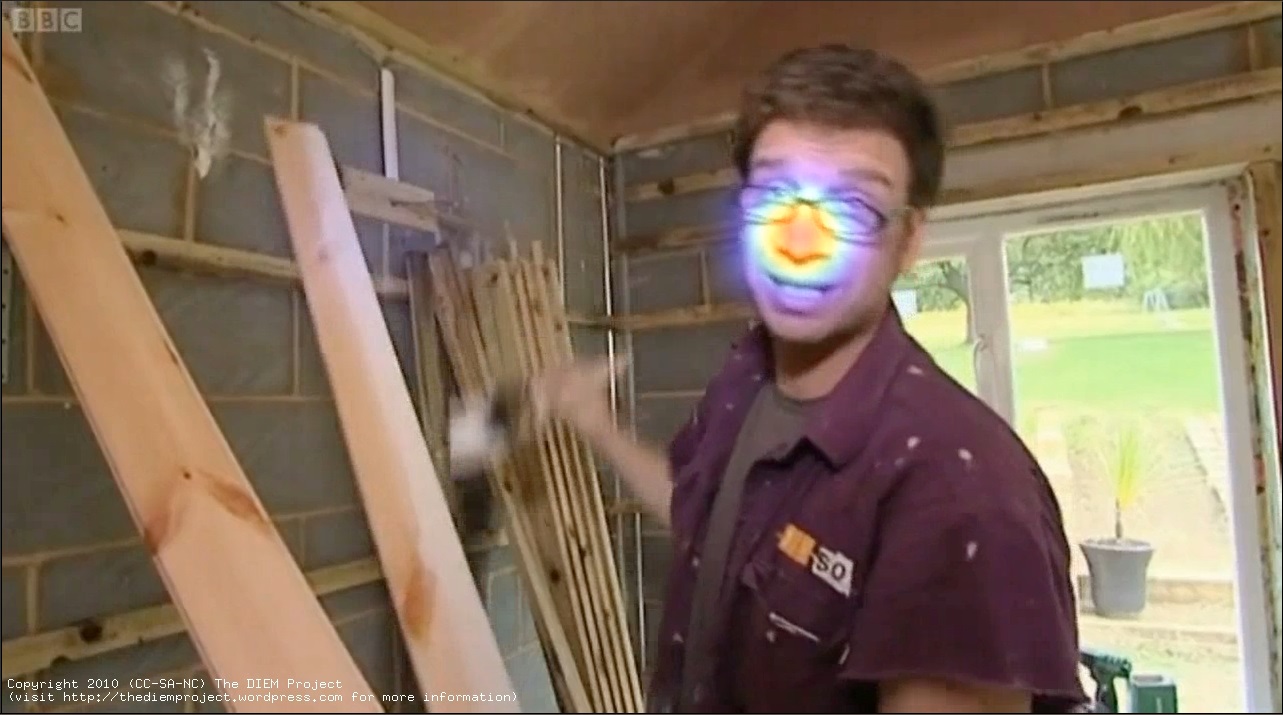}}    
    \hfill
    \subfloat{
    \includegraphics[width=0.3\textwidth]{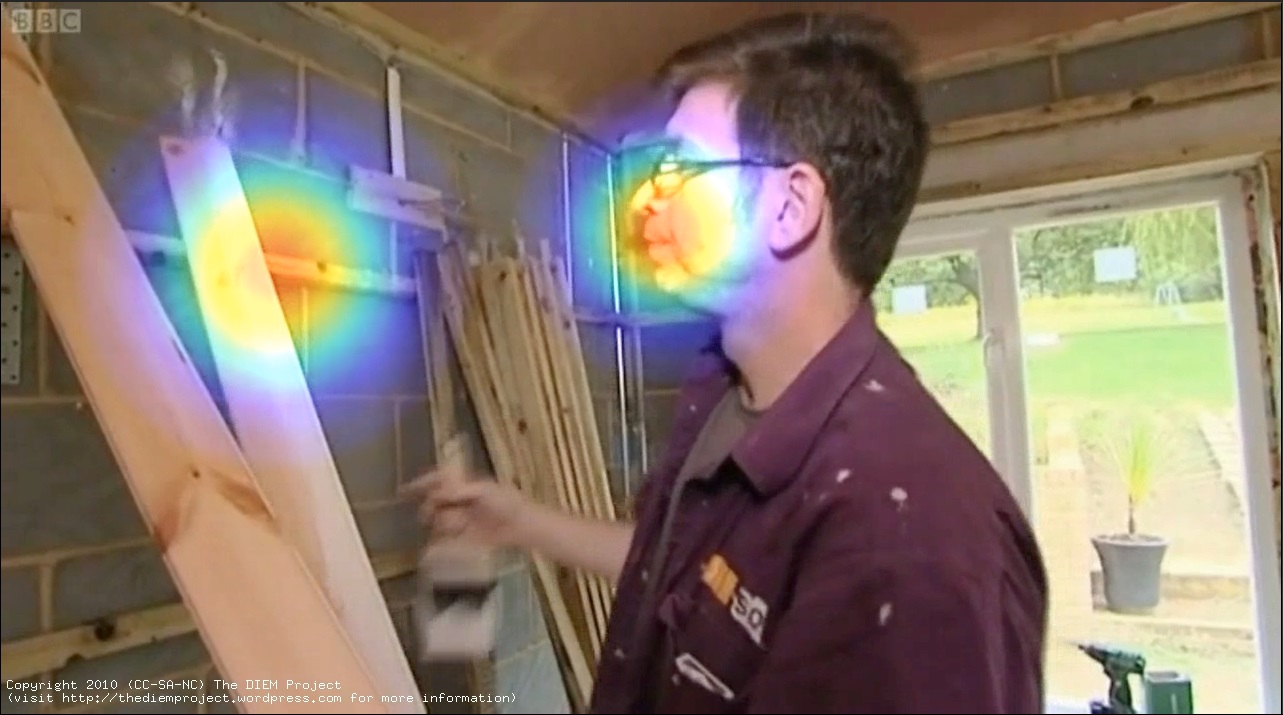}}            
	\hfill\null
\\[-1.7\baselineskip] 
	\centering
    \null\hfill
    \subfloat{
    \includegraphics[width=0.3\textwidth]{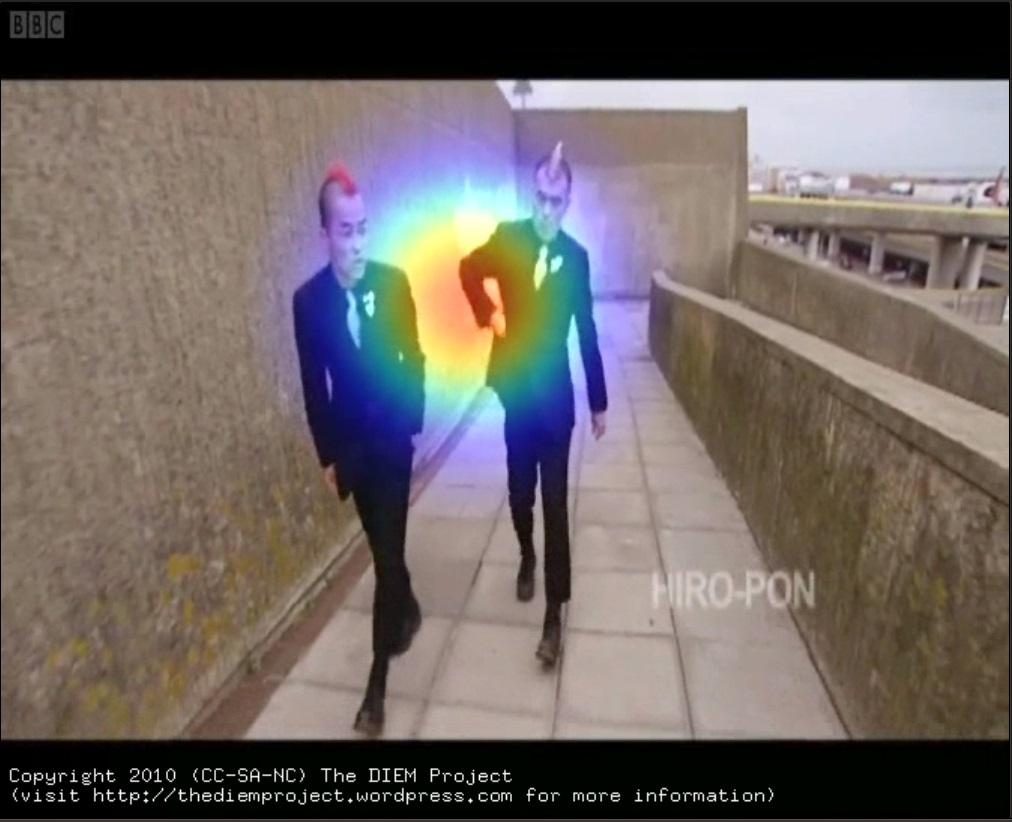}}       
    \hfill
    \subfloat{
    \includegraphics[width=0.3\textwidth]{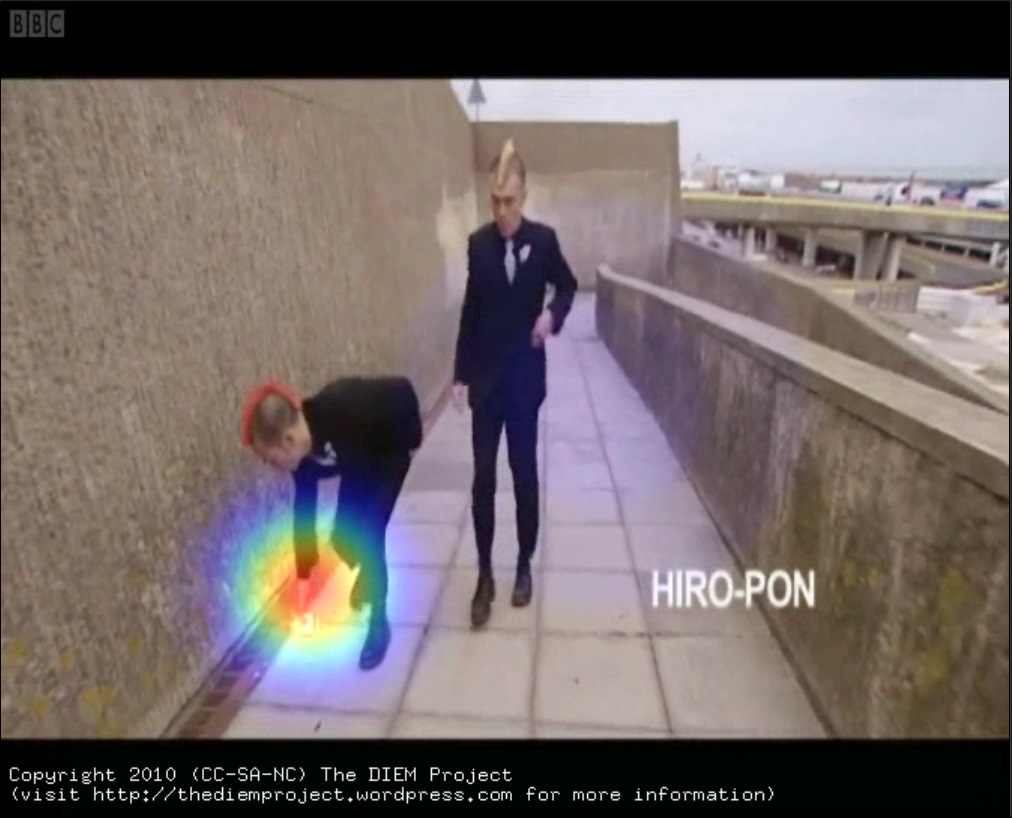}} 
    \hfill
    \subfloat{
    \includegraphics[width=0.3\textwidth]{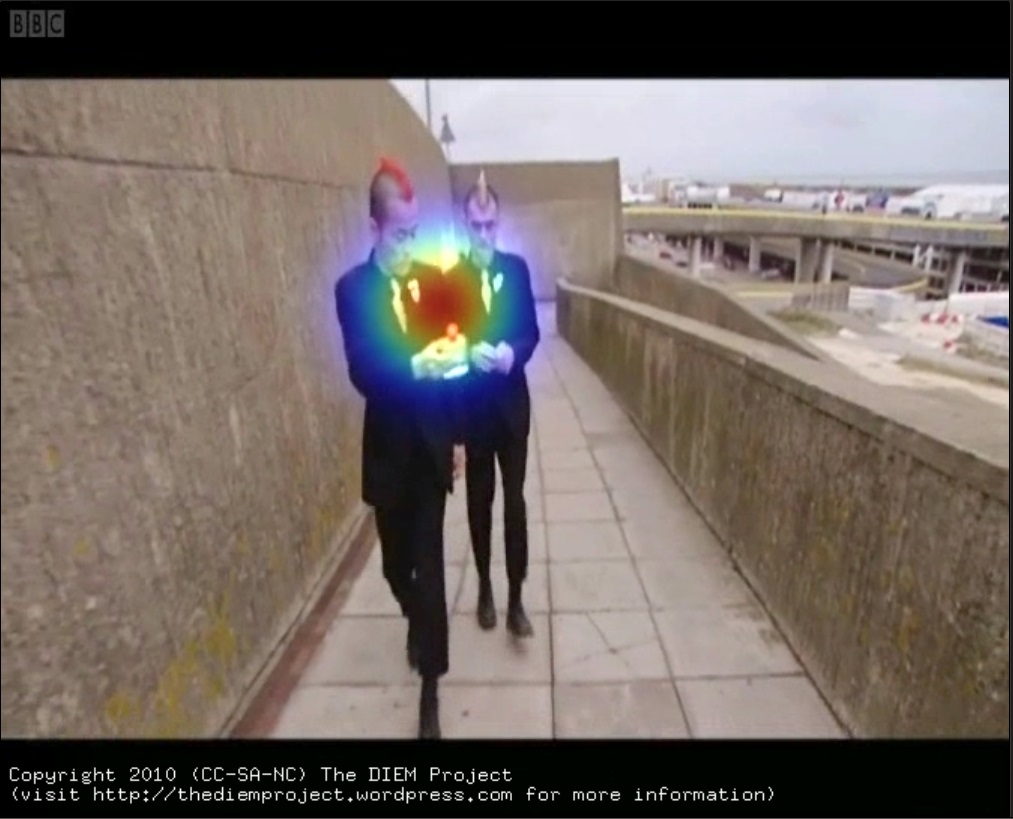}}   
	\hfill\null 
	\setlength{\belowcaptionskip}{-10pt}
    \caption{\label{fig:DynamicExamples}
Dynamic changes in gaze and body pose direction as Attentional Push cues. The images show overlaid fixation maps for three video frames, before, while and after a dynamic gaze/body change. In all cases, viewers' attention is highly influenced by the Attentional Push cues. Original video and eye fixation data are from the DIEM dataset \cite{DIEM}).}
\end{figure}

One of the shortcomings of the current approaches is that, for the most part, they concentrate on analyzing regions of the image for their power to attract attention. However, as noted above, in many instances, a region of the image may have low salience, but nonetheless still have attention allocated to it. 
Clearly, in such cases there are no salient features that attract attention to these regions. Instead, we propose that a viewer has their attention pushed to these regions by some high level process.
This suggests that in building an attention model we should go beyond image salience and instead of only computing the power of an image region to \textit{pull} attention to it, we should also consider the strength with which other regions of the image \textit{push} attention to the region in question.
 
Our proposed method models the viewer as a passive participant in the activity occurring in the scene. While the viewers cannot affect what is going on in the scene, their attentional state can nonetheless be influenced by the actors in the scene. We will treat every image viewing situation as one of \textit{Shared Attention}, which is the process by which multiple agents mutually estimate, direct and follow each others attentional state \cite{Kaplan2006}. 
As one of the building blocks for social communication, shared attention is a bilateral process by which an agent attends to an object that another agent attends to. 
Here, an agent may refer to both a scene actor and the viewer. To achieve shared attention, agents must observe, coordinate and influence their behaviors in order to engage in a collaborative intentional action \cite{Kaplan2006}.

We use the term \textit{Attentional Push} \cite{Smith2008} to refer to the power of image regions to direct and manipulate the attention allocation of the viewer.
Attentional Push can arise from many sources, which are mostly abstract high-level features, such as faces and body pose. 
For example in Fig. \ref{fig:Examples}, the head pose and the body pose of the scene actors manipulate the attention of the viewer. Such Attentional Push cues direct the viewers' attention to the shared locus of attention of the scene actors. Fig. \ref{fig:Examples} shows that although the shared loci of attention might have low salience, they have viewers' attention allocated to them nonetheless. It is also clear that two of the best-performing saliency methods (according to the MIT saliency benchmark \cite{MITbenchmark}), BMS \cite{Zhang2015} and eDN \cite{Vig2014}, perform poorly in predicting the fixation maps for such images with social clues. In addition, the strength by which an Attentional Push cue directs the viewers' attention could intensify as more actors focus their attention to the same shared locus of attention. 
 
We propose that the effect of Attentional Push in directing viewers' attention intensifies in more immersive scenarios, such as dynamic videos, 3-D movies and ultimately, while using virtual reality setups. Therefore, comparing to standard image salience-based methods, the prediction performance of an Attentional Push-based method would become more noticeable, as viewers feel more immersed in the ongoing event in the scene. Fig. \ref{fig:DynamicExamples} illustrates the effect of dynamic changes in gaze and body pose direction, as Attentional Push cues, on viewers' attention, while watching a dynamic movie. It suggests that as the level of immersion increased, viewers' attention is more influenced by Attentional Push cues.

This paper presents an attention tracking method that combines Attentional Push cues with standard image salience-based algorithms to improve the ability to predict where viewers' fixations in social scenes. Our approach to Shared Attention is to first identify the actors in the image, which can then be analyzed for their Attentional Push, potentially directing and manipulating the attention allocation of the viewer. 
The introduction of attention tracking and prediction techniques based on treating the viewer as a participant in a shared attention situation will open new avenues for research in the attention field. 

In a recent study, Parks et al. \cite{Parks2015} proposed the DWOC model, an attention model which combines bottom-up saliency with the head pose of the scene actors. The method is based on a two-state Markov chain describing the transition probabilities between head region and non-head region states, which is used to predict whether the next fixation is gaze related or being saliency driven. Our proposed method differs from Parks et al. \cite{Parks2015} in the following aspects: (i) their method only considers the effect of actors' head pose in manipulating the viewer's attention, whereas our Shared Attention-based method generalizes to all such Attentional Push cues; (ii) their method is only applicable to static scenes, whereas our method explicitly benefits from dynamic Attentional Push cues in directing viewers' attention while watching dynamic imagery; (iii) their method requires the viewers' eye movements to predict the next fixation point, whereas our method is based the image information only; and (iv) their method assumes the viewers have to fixate upon the head regions, in order for their next fixations to be influenced by the actors' gaze direction. However, this might not be the case and in our model the viewers' attention might be affected when the viewer tries to understand the gist of the scene.

The rest of this paper is organized as follows. Section \ref{Attentional Push} elaborates using Attentional Push in attention tracking. Section \ref{Augmented Saliency} presents our attention tracking model which augments standard saliency maps with Attentional Push cues. Section \ref{Evaluation and Comparison} illustrates experimental evaluation of the proposed method. Section \ref{Conclusion} concludes the paper.

\section{Attentional Push}
\label{Attentional Push}
To benefit from the Attentional Push cues in predicting viewers' attention, we propose to consider the viewer of the imagery as a partner in a shared attention situation, where the other partner(s) are the actors in the imagery. 
The goal of an agent in a shared attention setting is to coordinate its attention with other agents. To achieve this, the agent may try to interpret the intentions of another agent by watching its movements and its attentional behavior. 
While Kaplan and Hafner \cite{Kaplan2006} require the both agents to be able to detect, manipulate, coordinate and understand the attentional state and the behavior of the other agent in order to reach shared attention, our particular situation is a restricted asymmetric form of shared attention, in that the viewer has no control over the attentional state of the actors in the imagery. However, the actors in the image are assumed to have some control over the attentional state of the other actors in the image, as well as that of the viewer. 
Our working assumption will be that if two or more actors in a scene have a shared attentional locus, then the viewer will also be compelled to direct his or her attention to that locus. Thus, not only are we tracking the attention of the viewer, we are also tracking the attention of the actors in the scene, and doing so in a cooperative manner. 

Many Attentional Push cues have been reported in the literature of attention tracking. Perhaps the most prominent of these are gaze cues. Development of gaze following capabilities for robots via different learning mechanisms has been in the spotlight of research into socially interactive robots human-robot interaction (see the recent survey by Ferreira and Dias \cite{Ferreira2014} and the references therein). Castelhano et al. \cite{Castelhano2007} showed that while the actor's face is highly likely to be fixated, the viewer's next saccade is more likely to be toward the object that is fixated by the actor, compared to any other direction. Ricciardelli et al. \cite{Ricciardelli2002} showed that perceived gaze enhances attention if it is in agreement with the task direction, and inhibits it otherwise. They showed that in spite of top-down knowledge of its lack of usefulness, the perceived gaze automatically acts as an attentional cue and directs the viewer's attention. 
Similarly, as illustrated in Fig. \ref{fig:DynamicExamples}, the body pose of the scene actors could also push the viewers' attention. Although the attentional manipulation strength of the gaze direction dominates the body pose direction in most cases, it could be still intensified if the body pose direction is in agreement with the gaze direction. 

Apart from gaze and body pose cues, one of the most frequently cited Attentional Push cues in the literature is the center bias. Borji et al. \cite{Borji2013} showed that a simple 2D Gaussian shape drawn at the center of the image predicts the viewers' fixations well. We can treat the center-bias effect in the shared attention setting by considering the photographer as an actor in the shared attention setting, which tries to put the semantically interesting and therefore, salient elements in the center of the frame. In \cite{Tseng2009}, Tseng et al. showed that center bias is strongly correlated with photographer bias, rather than the viewing strategy and motor bias. There are some attention tracking models (e.g. Judd et al. \cite{Judd2009}) that have explicitly used the center-bias as a location prior to achieve better performance in predicting the eye movements. 

Aside from the static Attentional Push cues mentioned above, Attentional Push cues can also arise from dynamic events. For example, Smith \cite{Smith2012} showed that sudden movements of the heads of actors are a very strong cue for attention, where the viewer's FOA is not the head itself, but where it is pointing to (see Fig. \ref{fig:DynamicExamples}). Smith \cite{Smith2012} also notes the "bounce" in the attention of a movie viewer back to the center of the movie screen when tracking an object which moves off the screen to one side. Similarly, in \cite{Tseng2009}, abrupt scene changes are used to assess the contribution of the center bias in predicting viewer's attention while watching dynamic stimuli. 
We believe that employing such Attentional Push cues, either in static or in dynamic scenes, along with bottom-up image salience would be necessary to predict viewer's eye movements.

\section{Augmented Saliency}
\label{Augmented Saliency}
In this section, we present our attention tracking method which fuses the Attentional Push and the standard image salience techniques into a single attention tracking scheme. The proposed approach provides a framework for predicting viewer's FOA while watching static or dynamic imagery. 
For the sake of readability, the model focuses upon the interaction between one actor and the viewer, although this can be readily adapted in the case of multiple actors by providing unique identifiers for each actor. 
Our model distinguishes between two sets of attentional cues: Attentional Push-based and saliency-based, and provides a selection mechanism between them. While the saliency-based cues represent properties of the scene objects, the Attentional Push cues are based on the scene actor(s), such as head pose, body pose and dynamic changes in any of them as well as rapid scene changes. 
The need for a deterministic selection mechanism stems from the fact that in certain circumstances, an Attentional Push cue might pull the viewer's attention. An example of such situation is when a scene actor has frontal head directions. This traditional signal of Attentional Pull strictly pulls the viewer's attention to the actor rather than pushing it elsewhere. This has been exploited in many researches on gaze imitation and Shared Attention (e.g. see  \cite{Hoffman2006} and \cite{Moon2014}). In the top row of Fig. \ref{fig:DynamicExamples}, it could be seen that while the actor's head pose pushes the viewers' attention when the actor is looking sideways, it pulls the viewers' attention when the head pose is frontal. Therefore, it is vital to have a selection mechanism between pulling and pushing viewers' attention.

Assuming that the scene is observable via an image $I$, we can model the actor's attentional focus $A$ as conditionally dependent on the bottom-up factors such as location and appearance properties of the scene objects $\boldsymbol{O}=\{O_1,.,O_k\}$, as well as the top-down factors of the ongoing task of the scene, parameterized by $\textbf{T}$. We can then describe the attentional manipulation of the scene actors and the scene objects over the viewers' attention $V$ by employing a set of latent attentional cues $\{a_i\}$. In this Shared Attention setting, the attentional focus of the scene actors and the viewers are given by $P(A|\boldsymbol{O},\textbf{T})$ and $P(V|\{a_i\})$, respectively.
Learning and inferring the viewers' attention using the above dependencies requires the attentional foci of the scene actors. However, in most cases, the eye movements of the scene actors are not available. We hypothesize that this is not actually needed and we can directly employ some overt attentional measures of the actors, such as head gaze direction, body pose direction and hand gesture direction, to infer the viewer's attention.

\begin{figure}[t]
    \centering
    \includegraphics[width=0.5\linewidth]{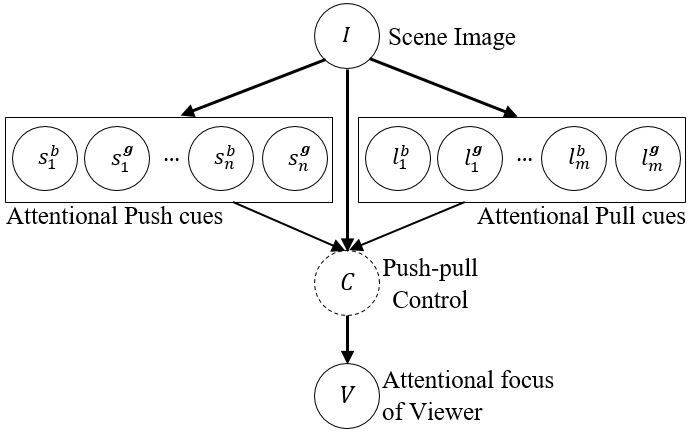}
    \label{fig:1a}
    \setlength{\belowcaptionskip}{-20pt}
    \caption{\label{fig:Model2}
    Shared Attention-based augmented saliency. The viewer's FOA $V$ is affected by the set of $n$ Attentional Push cues, represented by $\textbf{s}=\{s_i^b,s_i^\textbf{g}\}$, and the set of $m$ Attentional Pull cues, denoted by $\textbf{l}=\{l_i\}$. The model assumes the attentional cues to be directly observable via the scene image $I$. The deterministic node $C$ controls the transitions between the Attentional Push and Attentional Pull cues, based on their current observed values.}
\end{figure}

As shown in Figure \ref{fig:Model2}, we model the dependency between the attentional focus of the scene actors and the viewers by a set of $n$ observable Attentional Push cues $\textbf{s}=\{s_i^b,s_i^\textbf{g}\}$ and similarly, we use a set of Attentional Pull cues $\{l_i\}_{i=1}^m$, arising from image salience. The graphical model is used as a convenient method to describe the conditional dependencies of Attentional Push-based and saliency-based cues.
We employ normalized saliency maps $S(I)$ to estimate the joint distribution over the set of Attentional Pull cues $P(l_1,...,l_m,l|I)$. We represent each Attentional Push cue using two distinct quantities: 1) a geometrical structure $\textbf{g}:\{x,y,\boldsymbol {\theta},r,\sigma\}$, describing the $(x,y)$ location, 3-D rotation angles $(\boldsymbol{\theta}=\{roll,pitch,yaw\})$ (for symmetrical Attentional Push cues, $\boldsymbol{\theta}$ is set to the frontal direction), scale $(\sigma)$ and confidence factor $(r)$; and 2) a variable $b$ representing the presence or absence of the cue. For static Attentional Push cues, $b\in[0,1]$, while for dynamic Attentional Push cues, we encode the habituation factor \cite{Triesch2006}, i.e. the strength or probability of the viewers' motor response to a certain stimulus, by $b(t):=b(0)e^{-\beta (t-t_0)}$, where $\beta$ denotes the decay rate, $t_0$ is the moment of occurrence in which $b(0)$ is set and $t$ denotes discretized frame time.

We encode the deterministic constraints of the attentional guidance in the push-pull control node $C$ in Figure \ref{fig:Model2}. This node's value is deterministically assigned by its parents, using a predefined set of rules. For each Attentional Push cue $s_i$, we construct a 2-D Attentional Push map $M(s_i)$, having the advantage of being directly comparable with saliency maps. 
For directional Attentional Push cues, i.e. head/body pose and dynamic changes in head/body pose, we represent a 2-D map, having 1s along the direction $\phi_i$ ($\phi_i$ denotes the projection of $\boldsymbol{\theta_i}$ on the image plane), modulated by a 1-D Gaussian function centered at each point with a standard deviation proportional to $\sigma_i$ in the direction perpendicular to $\phi_i$ by $N(s_i)$. 
For symmetrical Attentional Push cues, i.e. frontal head pose, center bias, attentional bounce and rapid scene changes, we denote by $G(s_i)$ a 2-D map, containing a symmetric 2-D Gaussian, centered at the center of the 2-D map, with unit variance. 
The control node computes the Attentional Push maps by combining the directional and the symmetrical maps as follows: 
\begin{equation}
\label{eq:1}
M(s_i) = b(t)[\alpha \sigma_i G(s_i) + (1-\alpha)N(s_i)].
\end{equation}
where $\alpha$ is 1, if $\boldsymbol{\theta_i}$ is near frontal and 0 otherwise.

We employ the fusion mechanism in \cite{Marat2013} to combine the Attentional Push and Attentional Pull cues by assigning deterministic weights to each of them using their relevant statistics. For Attentional Pull cues $\textbf{l}=\{l_i\}$, we use the mean absolute skewness $\gamma$, i.e. the average of the absolute value of the third moments, of the normalized saliency map and for each Attentional Push map $\{M(s_i)\}$, we use its confidence measure $r_i$ in computing the weights. The control node output is  determined by
\begin{equation}
\label{eq:2}
C(\textbf{s},\textbf{l},I) = \gamma S(I) + \sum_{i=1}^{n}{r_iM(s_i)} + \gamma S(I)\sum_{i=1}^{n}{r_iM(s_i)}.
\end{equation}
Note that the third term in \ref{eq:2}, the element-wise multiplication of the saliency map and each Attentional Push map, acknowledges the fact that the directional Attentional Push maps are not able to discern between any image regions in the pose direction. The element-wise multiplication enables the directional Attentional Push-based cues and the saliency-based cues to interact in a way that if both of them have large values on a region, that region would have high saliency in the augmented saliency map.  

\section{Evaluation and Comparison}
\label{Evaluation and Comparison}
\subsection{Estimating Attentional cues}
To evaluate the performance of the Attentional Push-based method in predicting viewers' fixations, we employ the following Attentional Push cues: actors' body and head pose, the central bias, changes in actors' head and body pose, the bounce of attention and rapid scene changes. 
To identify the scene actors, we proceed by detecting humans and faces in the scene.
To detect humans, we employ the HoG-based detector of Dalal and Triggs \cite{Dalal2005}. To detect faces, we use the face detection system of Viola and Jones \cite{ViolaJones} and deformable mixture of parts-based method of Zhu and Ramanan \cite{ZhuRamanan2012}. Our experiments showed that the combination of the above methods results in a better detection rate, while increasing the false positive rate.  
For dynamic scenes, the scene actors might have non-frontal head poses which causes most face detection algorithms to fail. Therefore, we employ the state-of-the-art tracker TLD \cite{Kalal2012}, comprising of a median flow-based tracker, a detector, to localize the appearance of the faces, and a learning component which estimates the detector's error and updates it. The method returns a bounding box, computed from the merged results of the tracker and the detector. If neither the tracker nor the detector return a bounding box, the face is declared as non-visible which triggers a bounce of attention cue.
To estimate the head pose of the scene actors and their dynamic changes, we employ facial landmarks detection algorithms to accurately estimate the roll, pitch and yaw angles of the actor's head. Here, we use the iterative approach of \cite{Asthana2014} which initializes the landmarks locations using the face bounding box and uses an incremental cascaded linear regression to update the landmarks locations. 
To estimate the body pose direction, we use the poselet-based maethod of Maji, Bourdev and Malik \cite{Maji2011}.
To detect rapid scene changes, we adopt the method in \cite{Kim2000} which is based on comparing the edge strength and orientation of consecutive video frames.

\subsection{Evaluation protocol}
Attention models have commonly been validated against eye movements of human observers. To evaluate the proposed method, we employed three popular image and video datasets: 1) The CAT2000 dataset \cite{BorjiI15}, 2) the NUSEF dataset \cite{NUSEF}, and 3) The DIEM dataset \cite{DIEM}, containing eye movement data from 250 subjects watching 85 different dynamic scenes such as movie trailers, sport events and advertisements. 
Since the proposed Attentional Push-based method requires actors' in the scene, for the static stimuli, we used all the available images from the Action and the Social categories of the CAT2000 dataset (200 images in total). We also use 150 images from the NUSEF dataset. The employed images (350 images in total) contain humans and faces with resolution high enough for successful detection and accurate pose estimation. Note that if we run the proposed method for images with no actors, the results would be the same as the employed saliency method. For the dynamic stimuli, we use 13 videos from the DIEM dataset that contain people interacting with each other, each containing more than 1000 video frames (14109 video frames in total).
We compare our Attentional Push-based augmented saliency method with the ten best-performing state-of-the-art saliency models, according to the MIT saliency benchmark \cite{MITbenchmark} (see Table \ref{table:comparison}). For each saliency method, we create an augmented saliency using the proposed methodology. To evaluate attention models, many evaluation metrics have been proposed in the literature (e.g. \cite{MITbenchmark,Borji2013}). However, the performance of a model may change remarkably when different metrics are used. To ensure that the main qualitative conclusions are independent of the choice of metric, we analyze the performance of the proposed model using three popular evaluation metrics: the Area Under the ROC Curve ($AUC$), the Normalized Scan-path Saliency ($NSS$), and the Correlation Coefficient ($CC$). To compute $AUC$, fixated points are considered as the positive set while other locations are randomly sampled to form a negative set. By applying  multiple thresholds, the saliency map is used as a binary classifier and its ROC curve is plotted as the true positive rate against the false positive rate. Perfect prediction leads to an $AUC$ value of 1.0, while random prediction has an AUC of 0.5. The $NSS$ metric uses the average value of the saliency map, normalized to zero mean and unit variance, at fixation locations. When $NSS\geqslant1$, the saliency map exhibits significantly higher saliency values at human fixated locations compared to other locations. The $CC$ metric measures the strength of a linear relationship between the saliency map and the fixation map. Value of $abs(CC)$ close to 1 show a perfect linear relationship.

\begin{table}[t]
\centering
\caption{Average evaluation scores for the Attentional Push-based augmented saliency vs. ten best-performing saliency models on static and dynamic stimuli. The best performing method is shown in bold for each metric.}
\label{table:comparison}
\begin{tabular}{lcccccc}
\toprule 
• & \multicolumn{2}{c}{AUC} & \multicolumn{2}{c}{NSS} & \multicolumn{2}{c}{CC}\\

    & static & dynamic 
    & static & dynamic 
    & static & dynamic  \\
    \midrule

    AWS \cite{Garcia2012}   	    & 0.78 & 0.79 & 1.16 & 1.02	& 0.31 & 0.16	\\
    augmented AWS   	            & 0.85 & 0.91 & 1.66 & \textbf{2.44}	& 0.44 & \textbf{0.37}	\\ \hdashline
    
	BMS \cite{Zhang2015}   	& 0.80 & 0.80 & 1.19 & 1.15	& 0.31 & 0.17	\\
    augmented BMS   	    & 0.85 & 0.90 & 1.63 & 2.30	& 0.43 & 0.35	\\ \hdashline   
    
    Center \cite{MITbenchmark}   	& 0.61 & 0.75 & 0.47 & 0.99	& 0.13 & 0.15	\\
    augmented Center   	            & 0.77 & 0.90 & 1.20 & 2.26	& 0.32 & 0.35	\\\hdashline

    ContextAware \cite{Goferman2012}   	& 0.79 & 0.66 & 1.18 & 0.40	& 0.31 & 0.06	\\
    augmented ContextAware   	        & 0.85 & 0.88 & 1.61 & 2.10	& 0.43 & 0.31	\\ \hdashline   
    
    eDN \cite{Vig2014}   	& 0.85 & 0.90 & 1.23 & 1.43	& 0.33 & 0.22	\\
    augmented eDN   	    & \textbf{0.87} & \textbf{0.92} & 1.58 & 2.21	& 0.42 & 0.34	\\\hdashline
    
    FES \cite{Tavakoli2011} & 0.82 & 0.83 & 1.49 & 0.97	& 0.39 & 0.15	\\
    augmented FES   	    & 0.85 & 0.89 & \textbf{1.77} & 2.16	& \textbf{0.47} & 0.33	\\\hdashline
    
    GBVS \cite{Harel2007}   	    & 0.81 & 0.85 & 1.31 & 1.36	& 0.35 & 0.21	\\
    augmented GBVS   	            & 0.85 & 0.90 & 1.61 & 2.29	& 0.43 & 0.35	\\    \hdashline

    IttiKoch2 \cite{Itti98}   	    & 0.79 & 0.80 & 1.17 & 1.04	& 0.31 & 0.16	\\
    augmented IttiKoch2   	        & 0.85 & 0.90 & 1.59 & 2.13	& 0.42 & 0.33	\\        \hdashline
    
    Judd \cite{Judd2009}   	& 0.84 & 0.87 & 1.30 & 1.34	& 0.35 & 0.21	\\
    augmented Judd   	    & 0.86 & 0.91 & 1.61 & 2.21	& 0.43 & 0.34	\\\hdashline
    
    RARE \cite{Riche2013642}   	& 0.80 & 0.75 & 1.25 & 0.54	& 0.33 & 0.08	\\
    augmented RARE   	        & 0.85 & 0.89 & 1.66 & 2.16	& 0.44 & 0.33	\\ \hline            
    
    Average improvements	& 0.056	& 0.10	& 0.42	& 1.19	& 0.11	& 0.18	\\
               
\bottomrule
\end{tabular}
\end{table}

\subsection{Results and Discussion}
Table \ref{table:comparison} compares the prediction performance of the Attentional Push-based augmented saliency with the standard saliency methods for both static and dynamic stimuli.  
The results show that each of the augmented saliency methods improves its corresponding saliency method and the average evaluation scores for the augmented saliency methods are significantly higher than the average scores of the standard saliency methods.
For static stimuli, the most significant performance boost in AUC score is achieved by augmenting the AWS method (although the augmented Center model has the highest improvement, its AUC score is insignificant compared to the best performing method). The average performance boost over all of the augmented methods are 0.056, 0.42 and 0.11 for AUC, NSS and CC, respectively. It should be noted that the augmented saliency method is not only outperforming models that employ face and people detection such as Judd \cite{Judd2009}, it is also improving the prediction performance of data-driven methods such as the ensemble of Deep Networks (eDN) \cite{Vig2014}. 

\begin{figure}[]
\captionsetup[subfigure]{labelformat=empty} 
    \centering
    \null\hfill
    {\tiny{(a)}}
    \hfill    
    \subfloat{
    \includegraphics[width=0.15\textwidth]{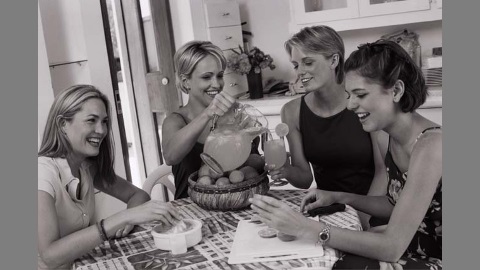}}    
    \hfill
    \subfloat{
    \includegraphics[width=0.15\textwidth]{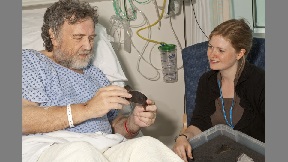}}
    \hfill
    \subfloat{
    \includegraphics[width=0.15\textwidth]{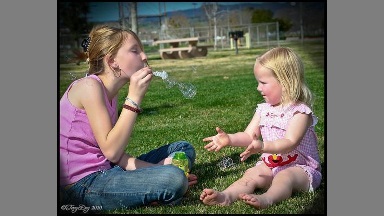}}
    \hfill
    \subfloat{
    \includegraphics[width=0.15\textwidth]{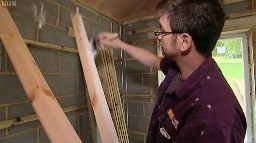}}
    \hfill
    \subfloat{
    \includegraphics[width=0.15\textwidth]{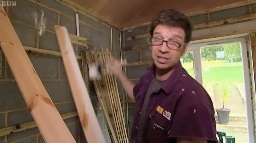}}
    \hfill
    \subfloat{
    \includegraphics[width=0.15\textwidth]{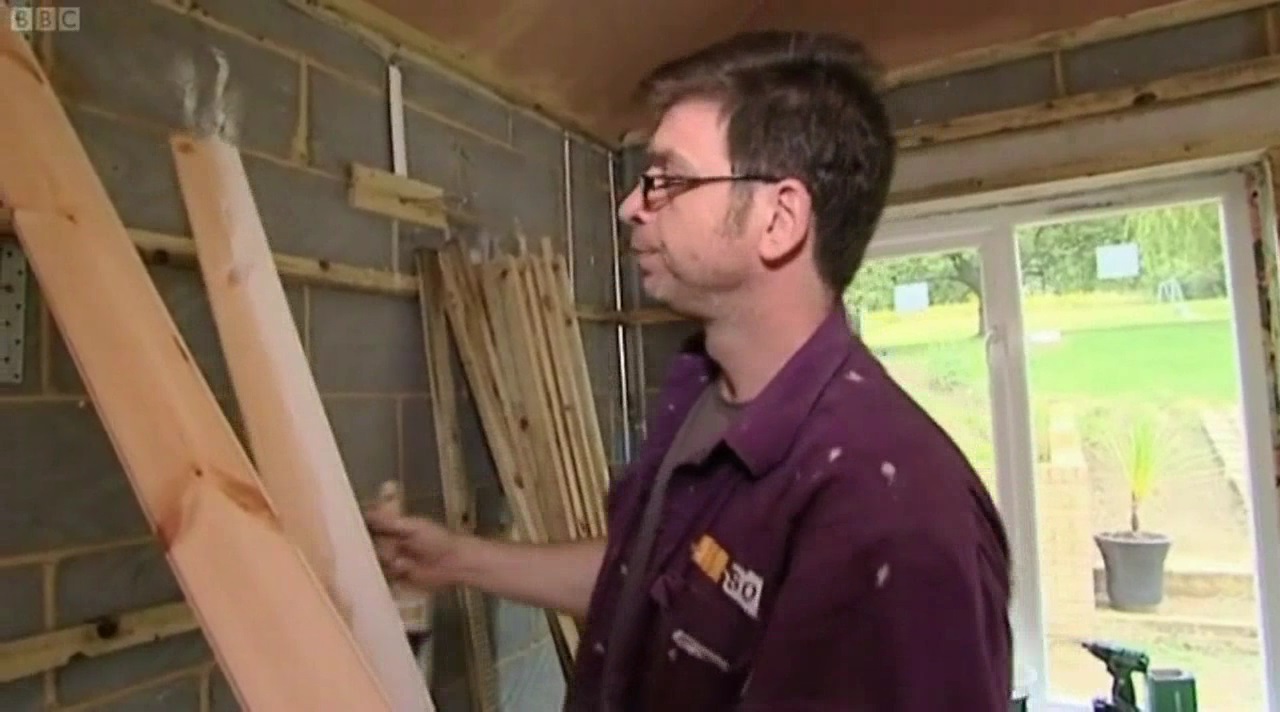}}
	\hfill\null
\\[-1.7\baselineskip] 
    \centering
    \null\hfill
    {\tiny{(b)}}
    \hfill
    \subfloat{
    \includegraphics[width=0.15\textwidth]{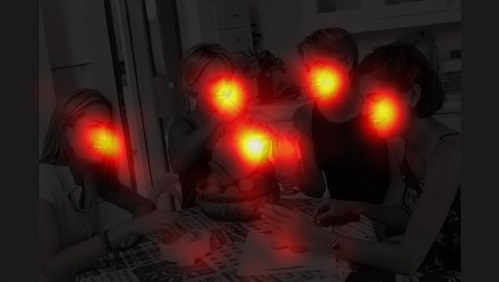}}    
    \hfill
    \subfloat{
    \includegraphics[width=0.15\textwidth]{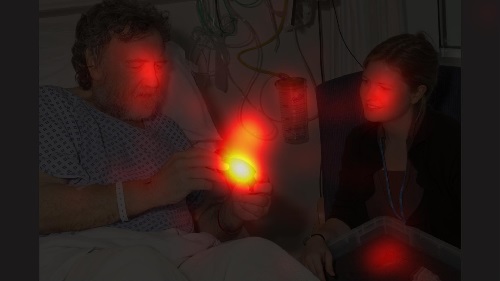}}
    \hfill
    \subfloat{
    \includegraphics[width=0.15\textwidth]{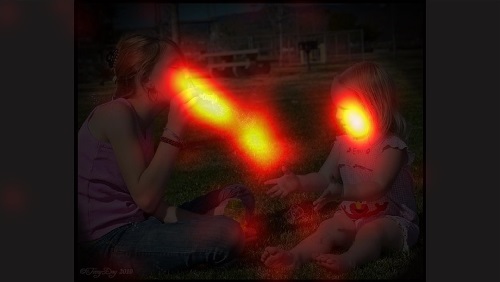}}
    \hfill
    \subfloat{
    \includegraphics[width=0.15\textwidth]{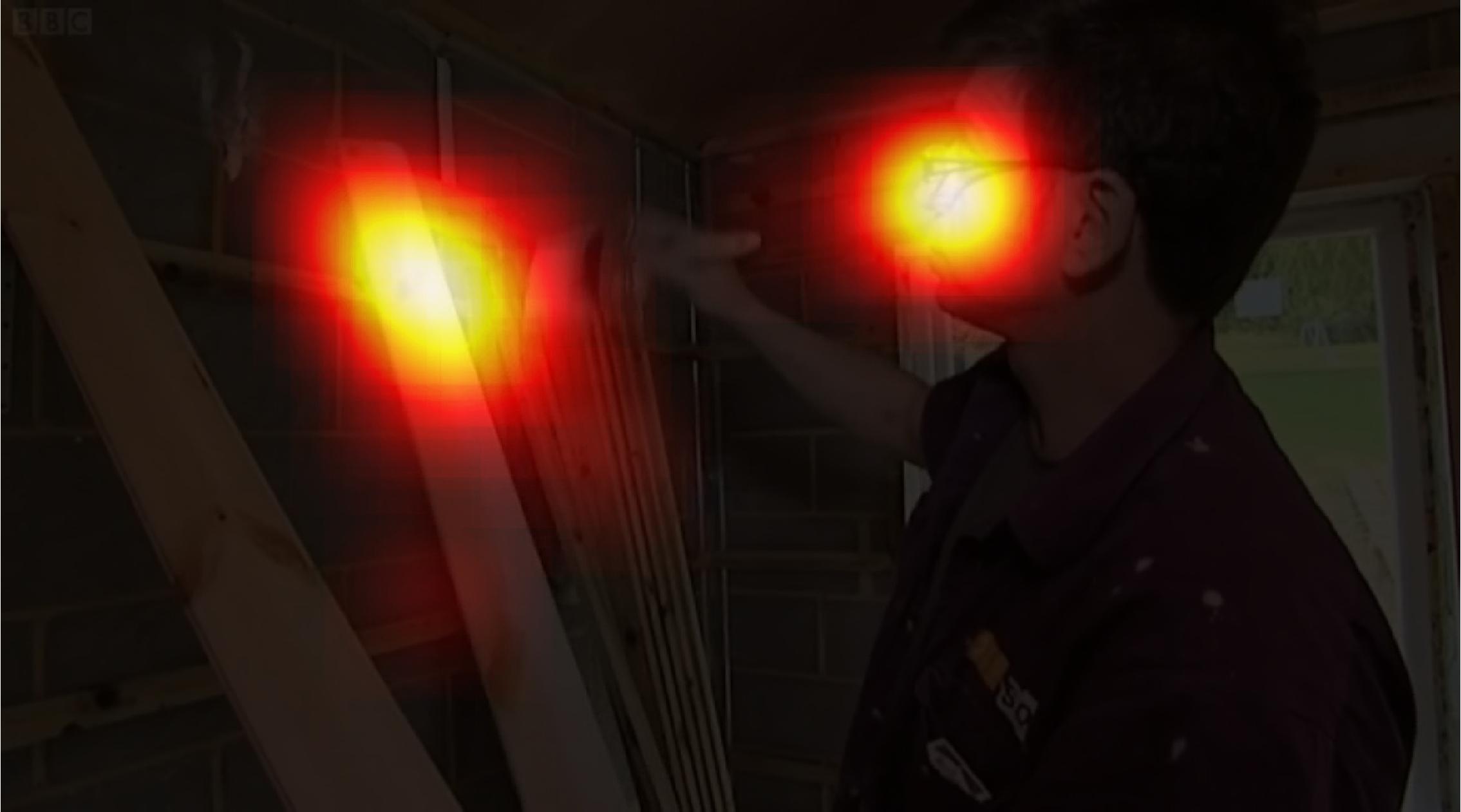}}
    \hfill
    \subfloat{
    \includegraphics[width=0.15\textwidth]{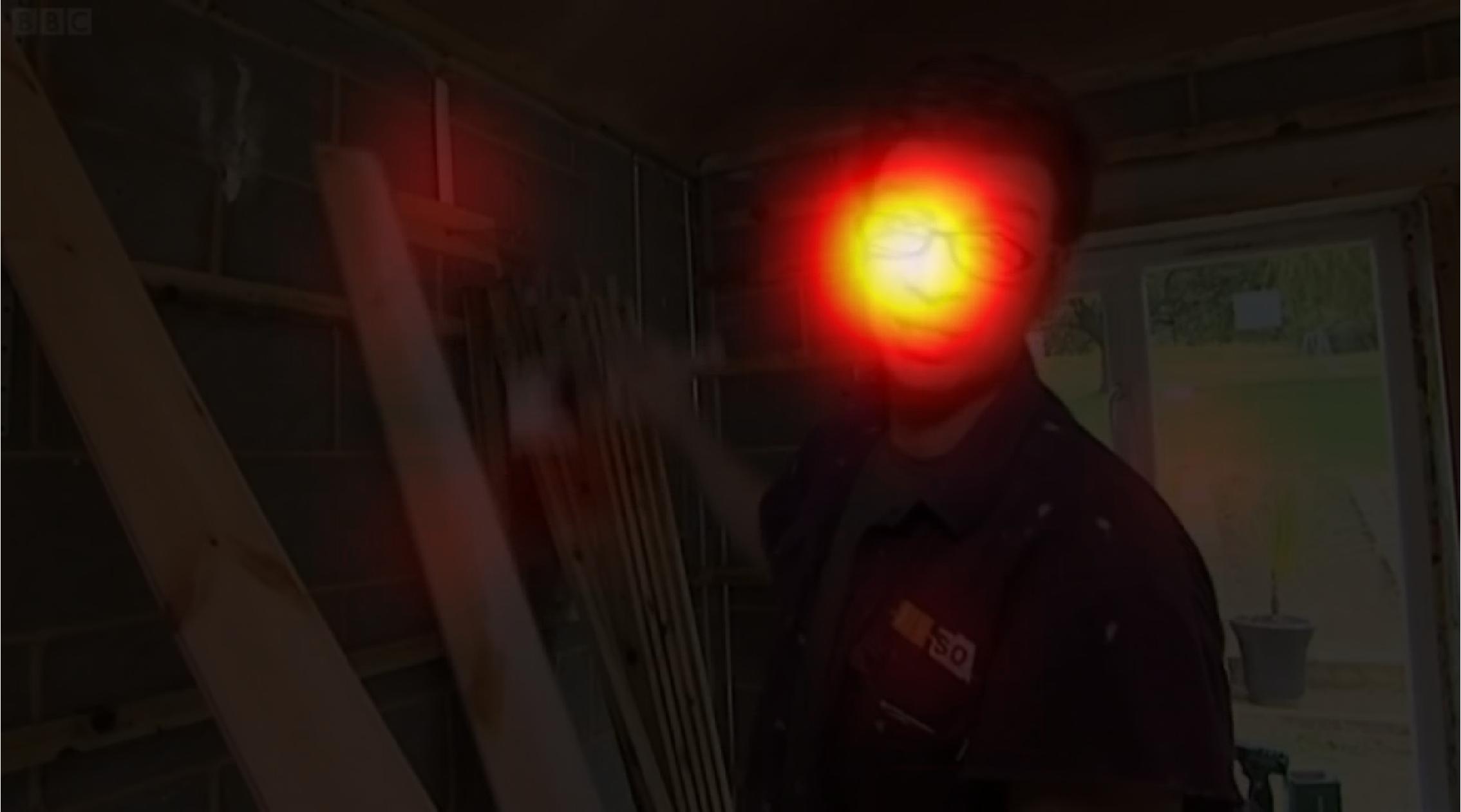}}
    \hfill
    \subfloat{
    \includegraphics[width=0.15\textwidth]{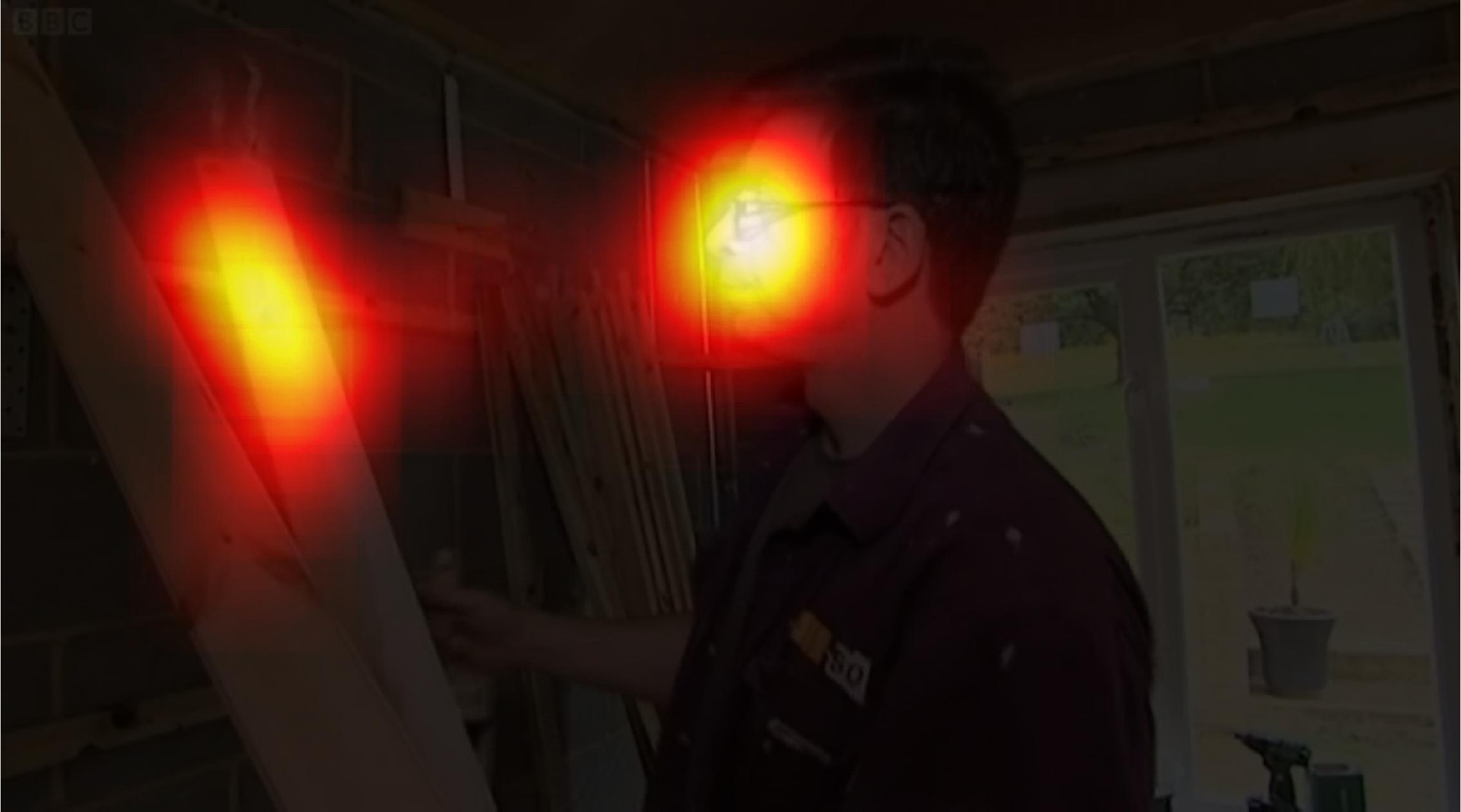}}
	\hfill\null
\\[-1.7\baselineskip]	
    \centering
    \null\hfill
    {\tiny{(c)}}
    \hfill
    \subfloat{
    \includegraphics[width=0.15\textwidth]{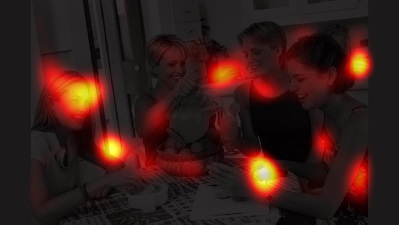}}    
    \hfill
    \subfloat{
    \includegraphics[width=0.15\textwidth]{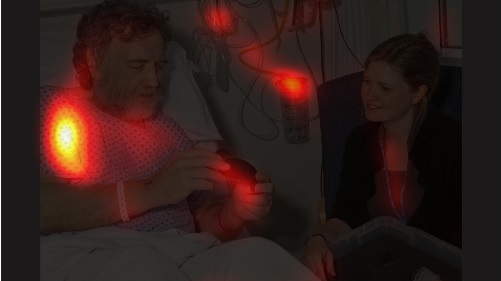}}
    \hfill
    \subfloat{
    \includegraphics[width=0.15\textwidth]{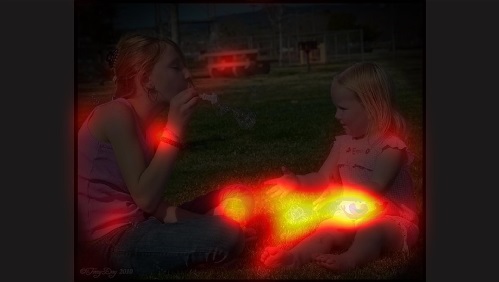}}
    \hfill
    \subfloat{
    \includegraphics[width=0.15\textwidth]{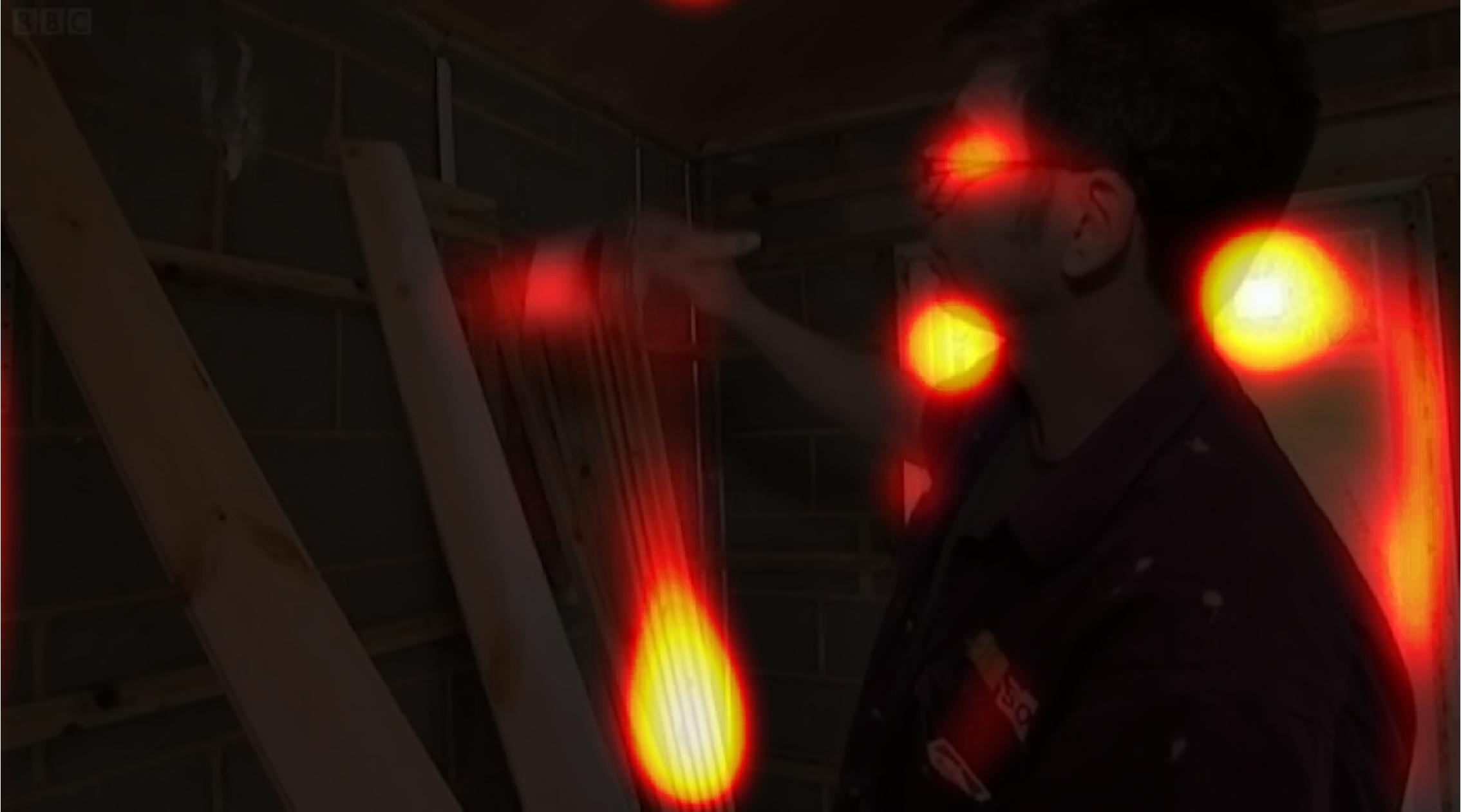}}
    \hfill
    \subfloat{
    \includegraphics[width=0.15\textwidth]{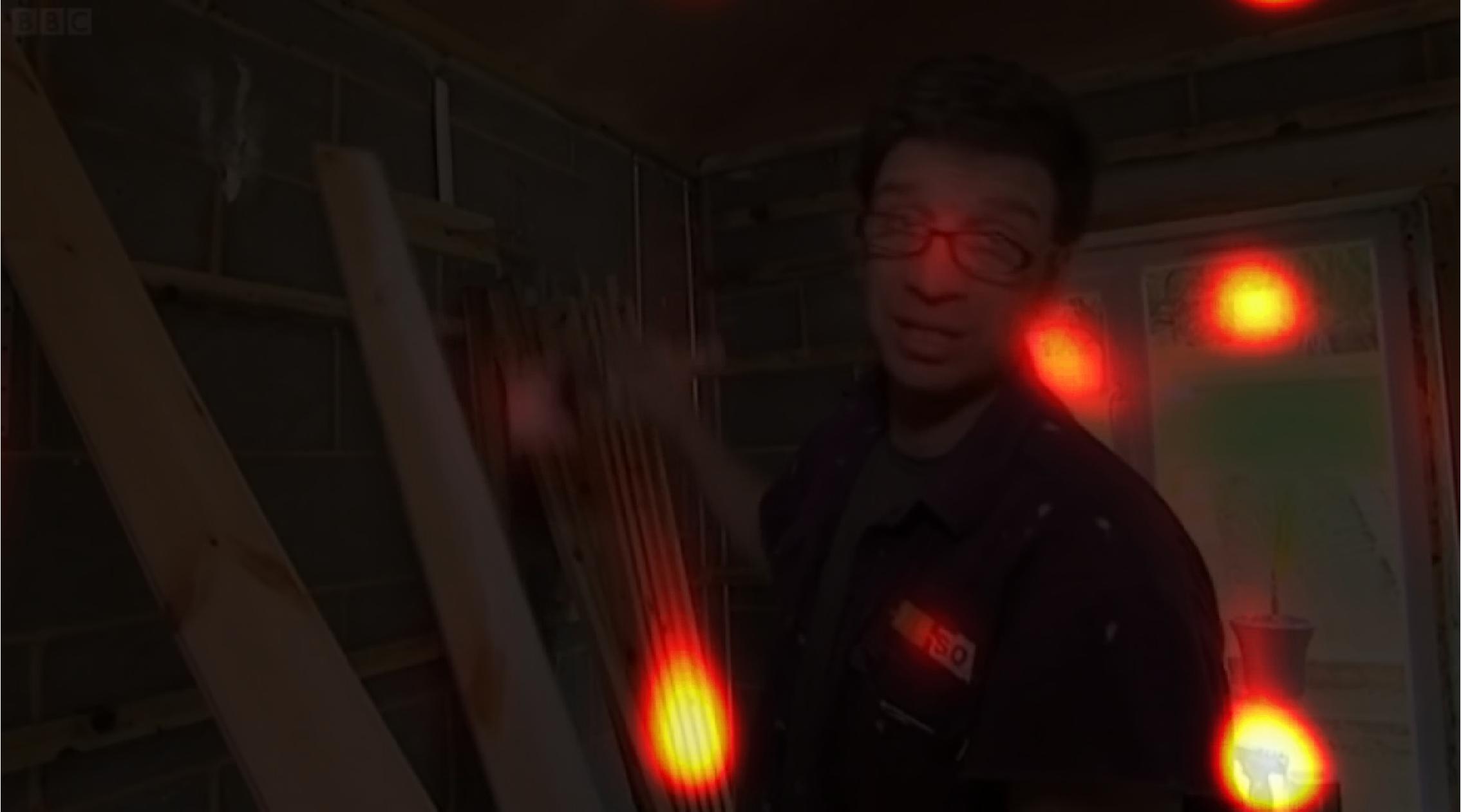}}
    \hfill
    \subfloat{
    \includegraphics[width=0.15\textwidth]{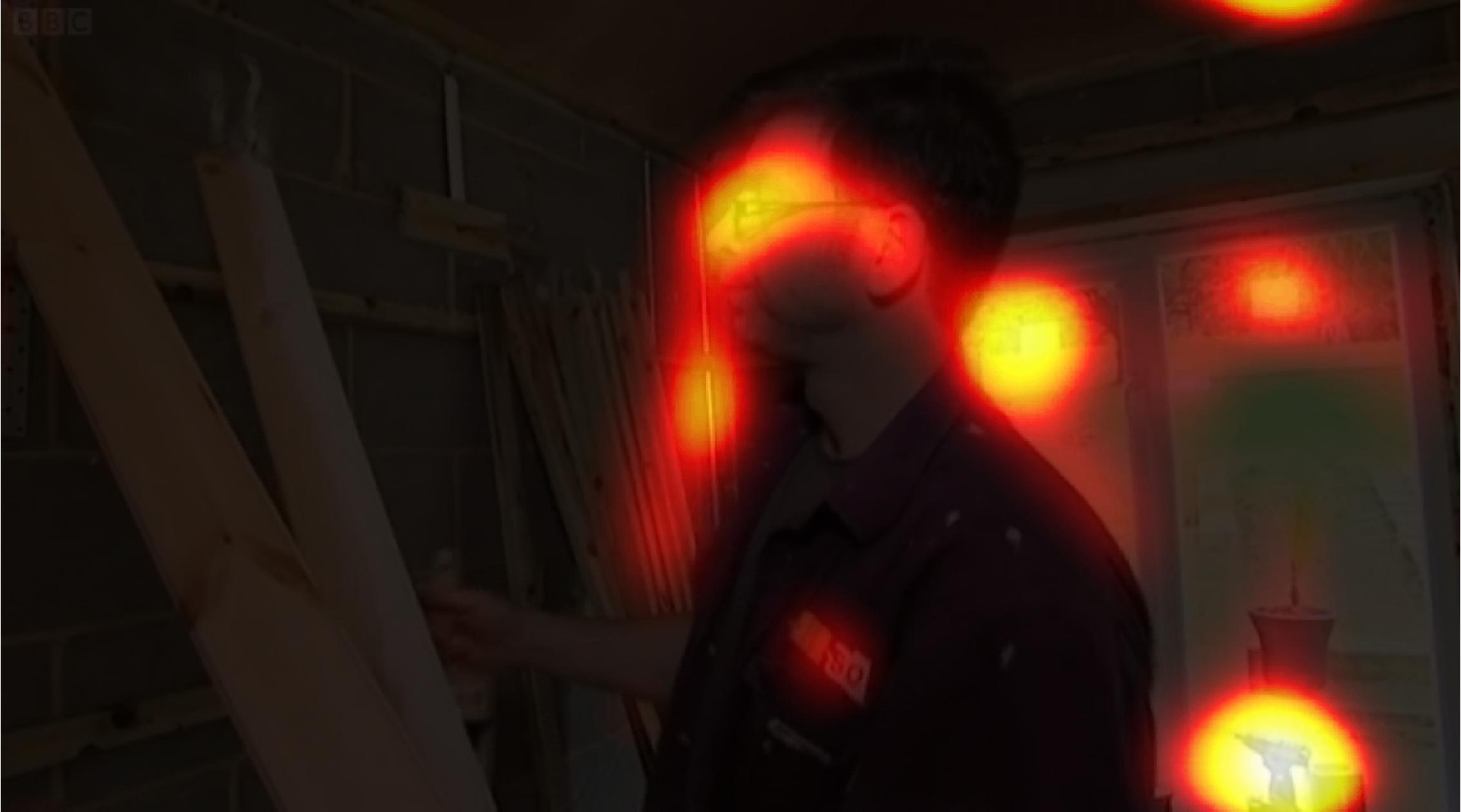}}
	\hfill\null
\\[-1.7\baselineskip]	
    \centering
    \null\hfill
    {\tiny{(d)}}
    \hfill
    \subfloat{
    \includegraphics[width=0.15\textwidth]{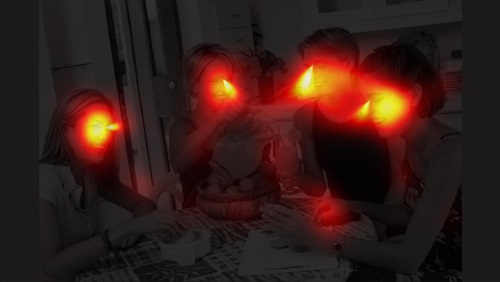}}    
    \hfill
    \subfloat{
    \includegraphics[width=0.15\textwidth]{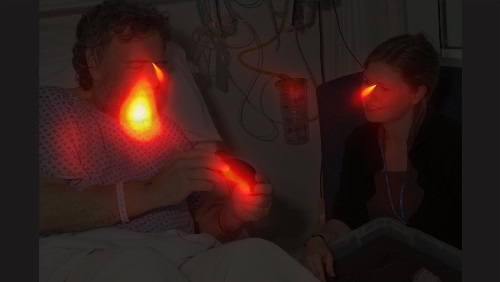}}
    \hfill
    \subfloat{
    \includegraphics[width=0.15\textwidth]{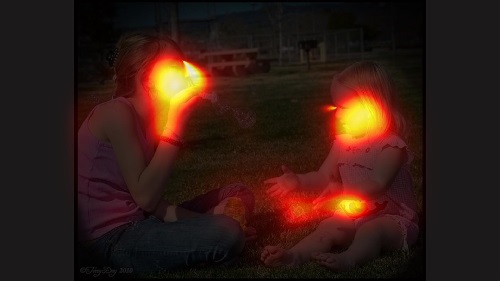}}
    \hfill
    \subfloat{
    \includegraphics[width=0.15\textwidth]{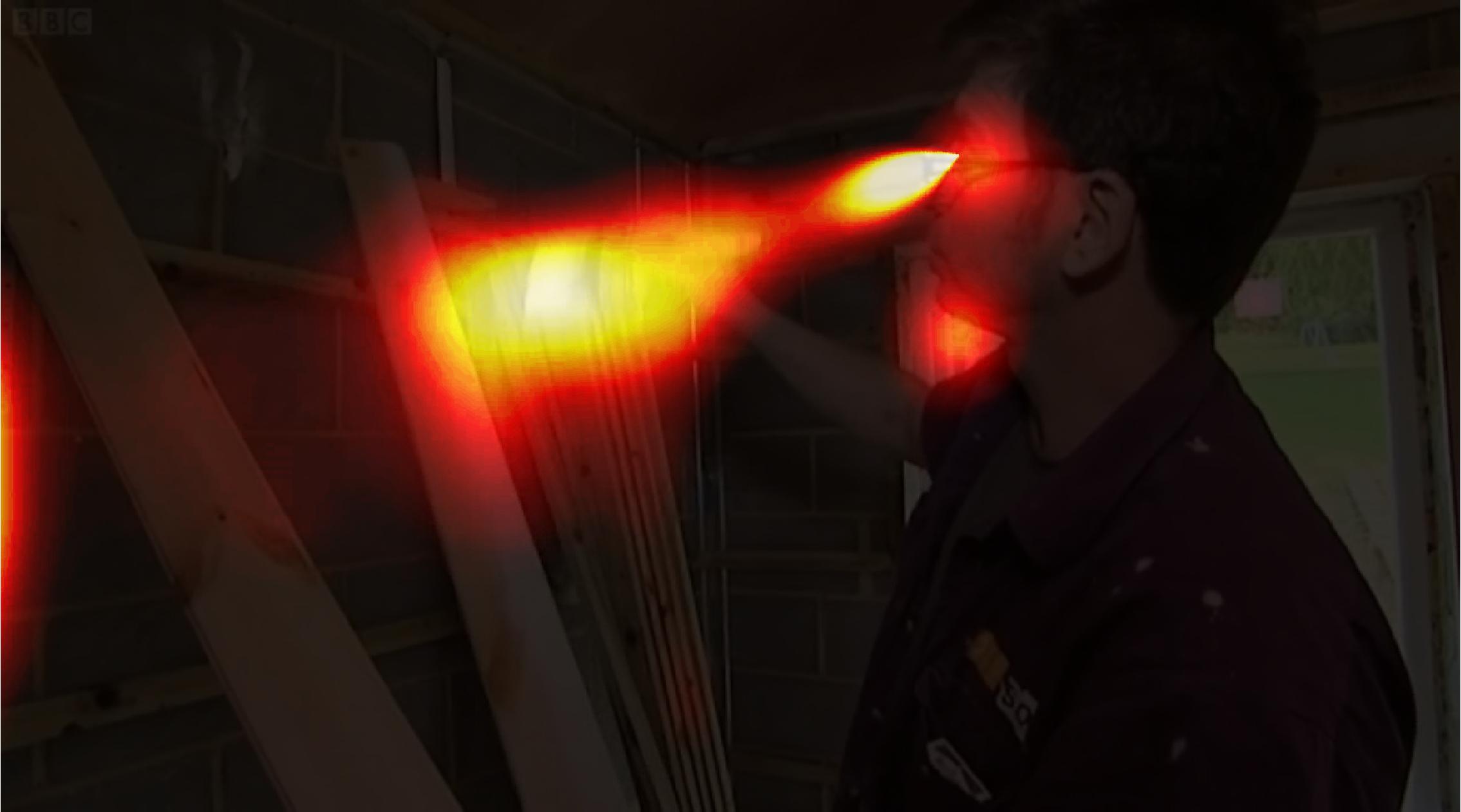}}
    \hfill
    \subfloat{
    \includegraphics[width=0.15\textwidth]{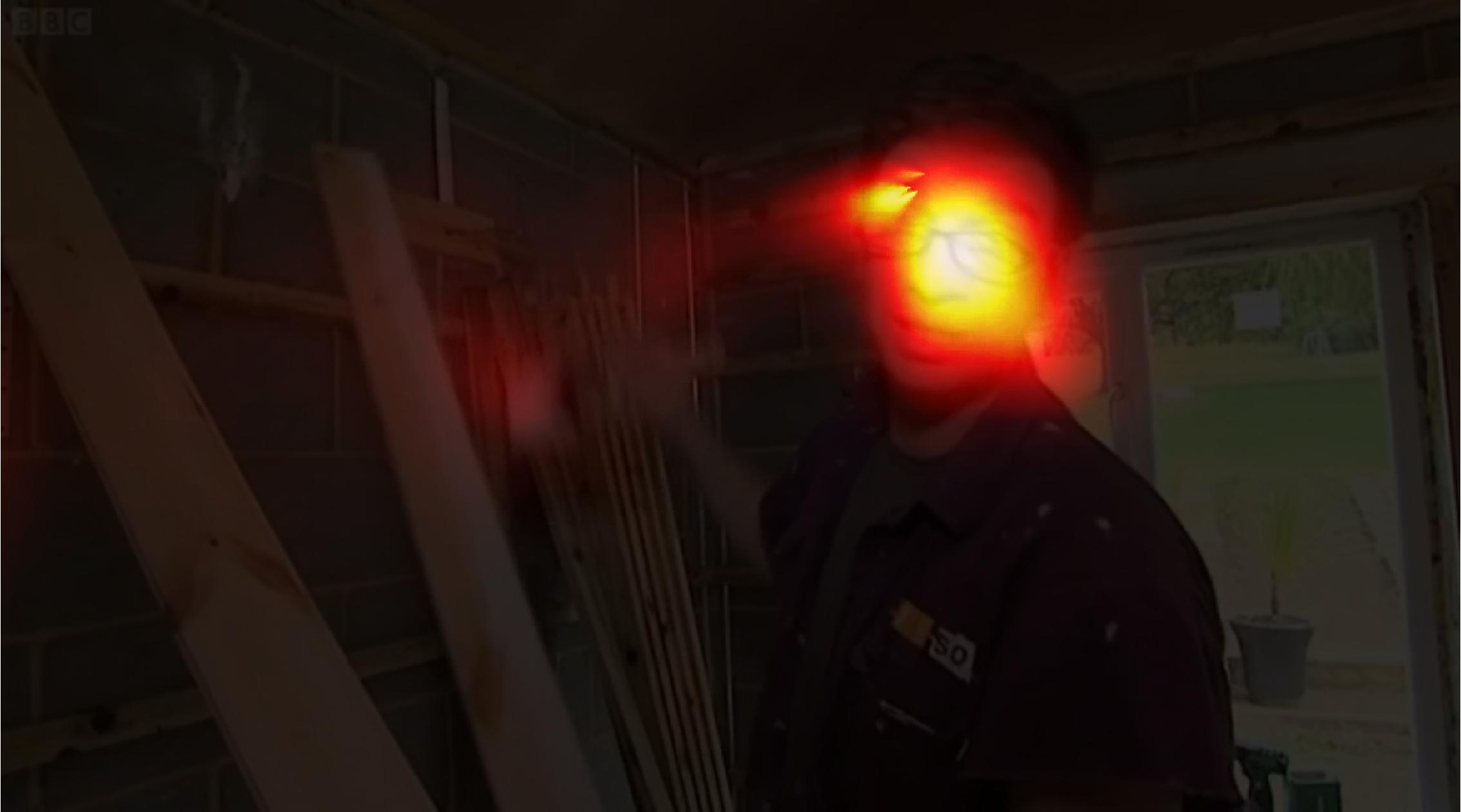}}
    \hfill
    \subfloat{
    \includegraphics[width=0.15\textwidth]{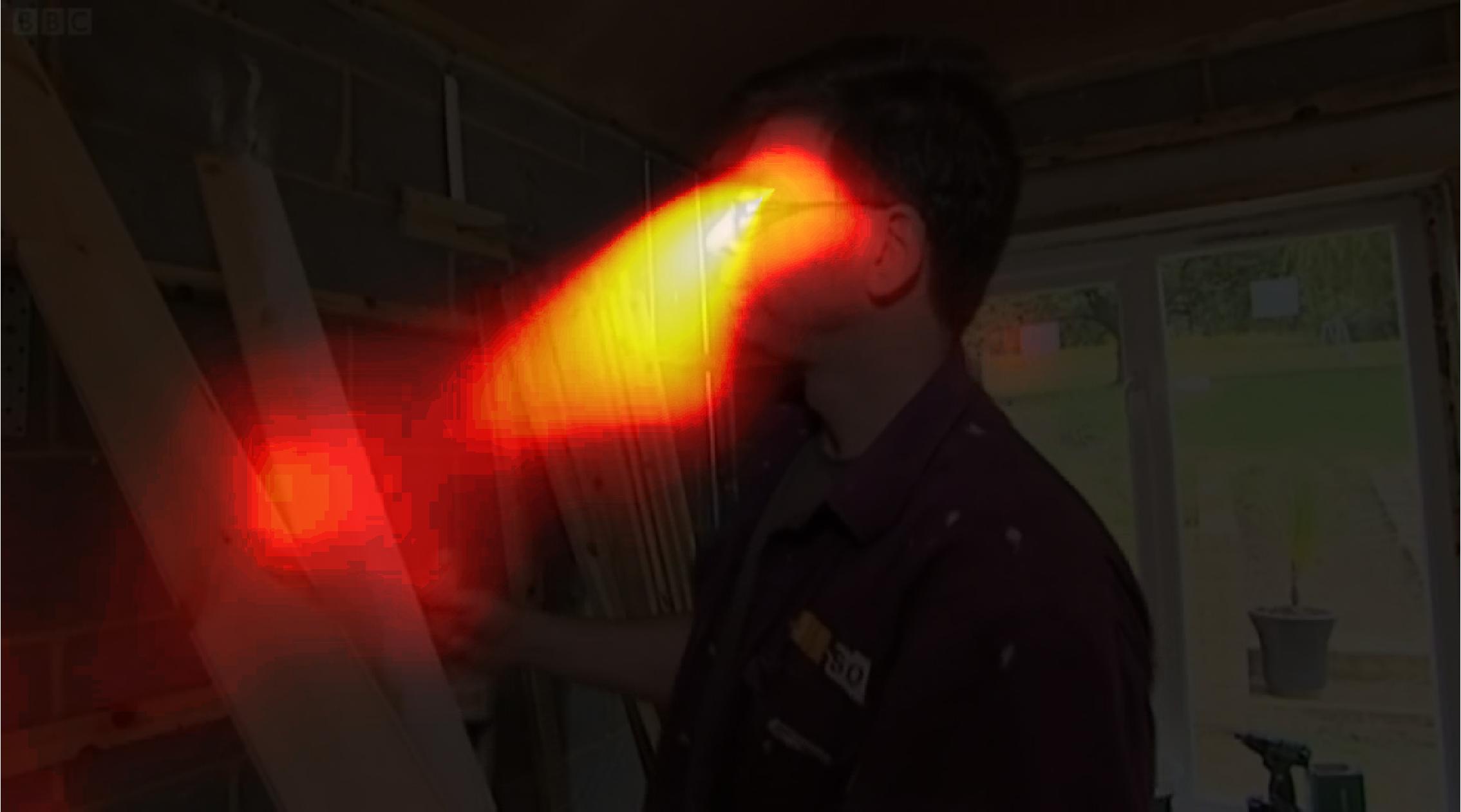}}
	\hfill\null	
\\[-1.7\baselineskip]	
    \centering
    \null\hfill
    {\tiny{(e)}}
    \hfill
    \subfloat{
    \includegraphics[width=0.15\textwidth]{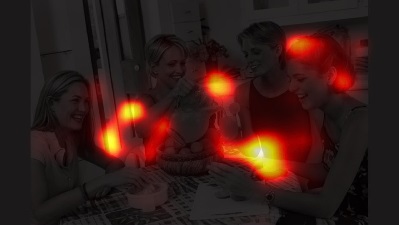}}    
    \hfill
    \subfloat{
    \includegraphics[width=0.15\textwidth]{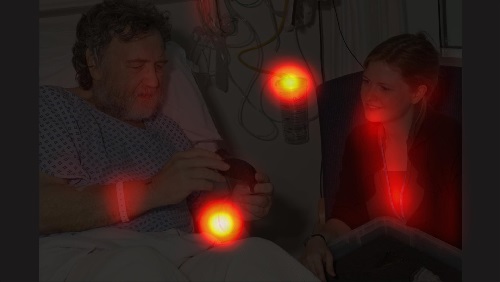}}
    \hfill
    \subfloat{
    \includegraphics[width=0.15\textwidth]{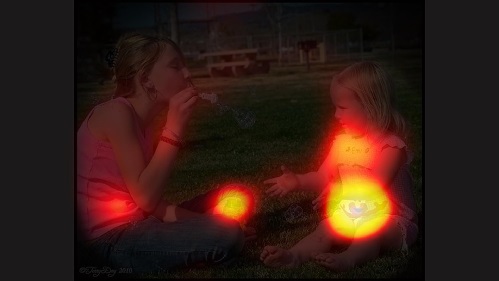}}
    \hfill
    \subfloat{
    \includegraphics[width=0.15\textwidth]{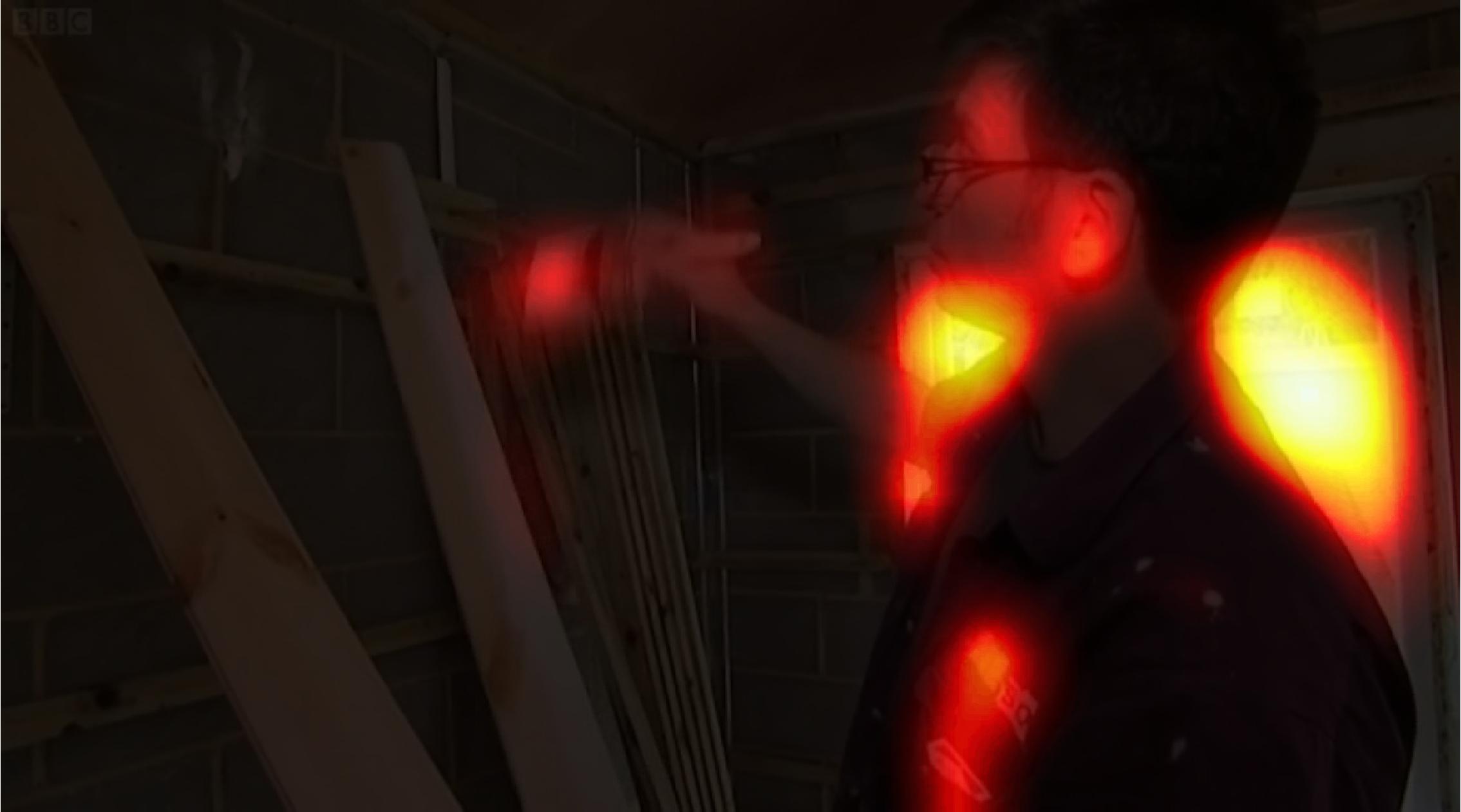}}
    \hfill
    \subfloat{
    \includegraphics[width=0.15\textwidth]{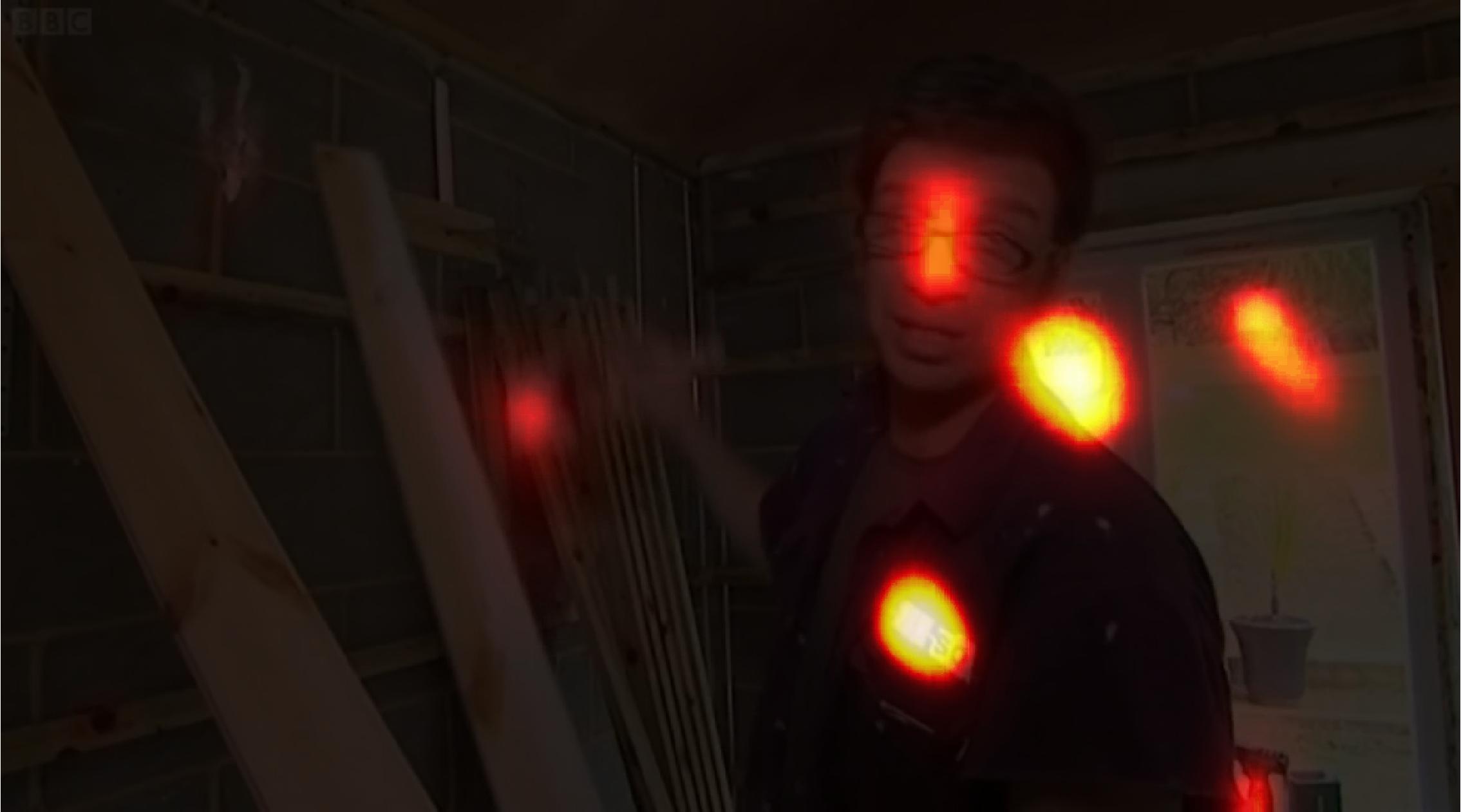}}
    \hfill
    \subfloat{
    \includegraphics[width=0.15\textwidth]{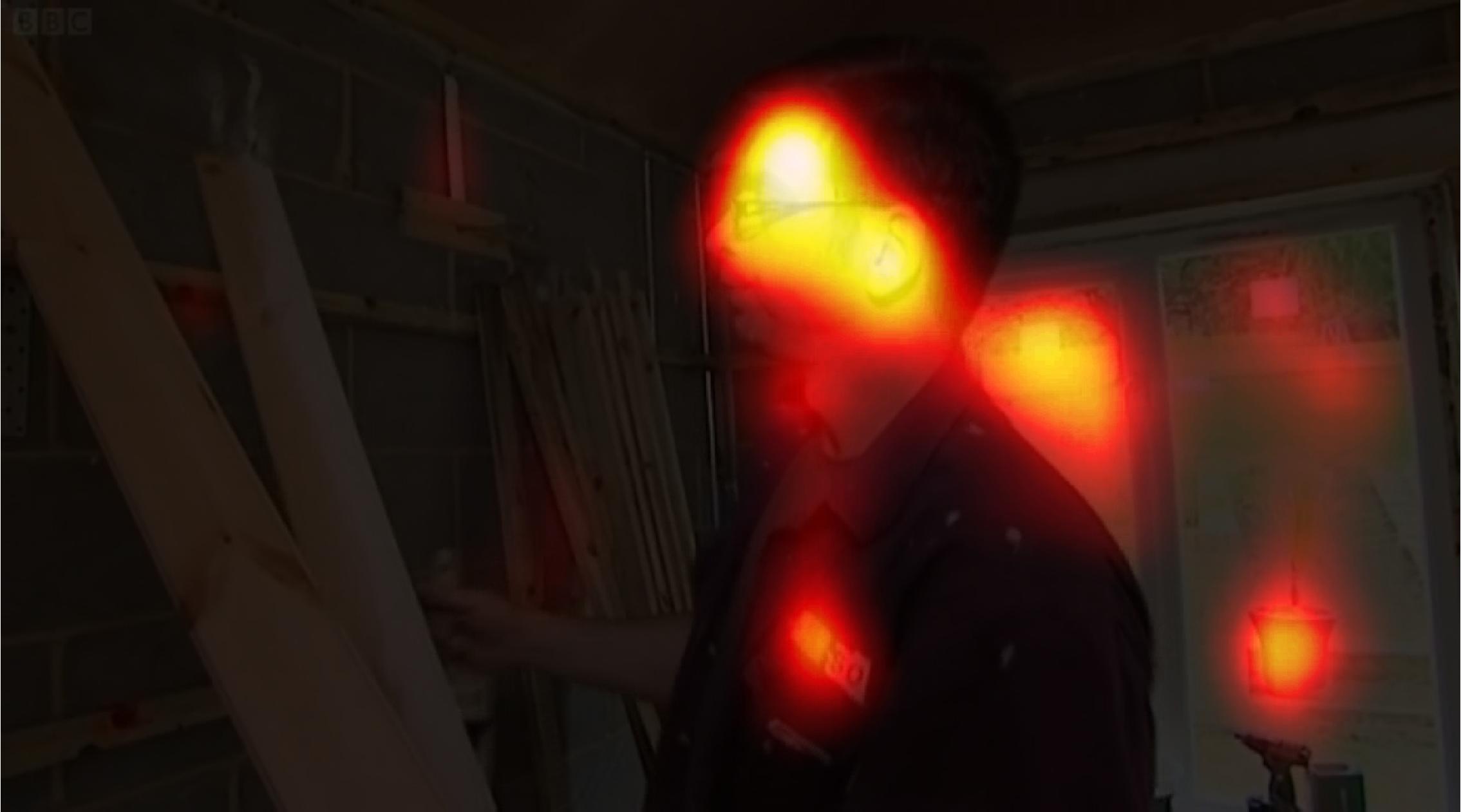}}
	\hfill\null
\\[-1.7\baselineskip]	
    \centering
    \null\hfill
    {\tiny{(f)}}
    \hfill
    \subfloat{
    \includegraphics[width=0.15\textwidth]{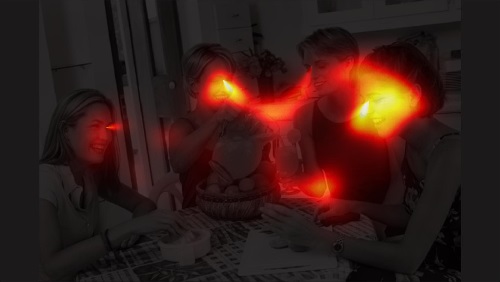}}    
    \hfill
    \subfloat{
    \includegraphics[width=0.15\textwidth]{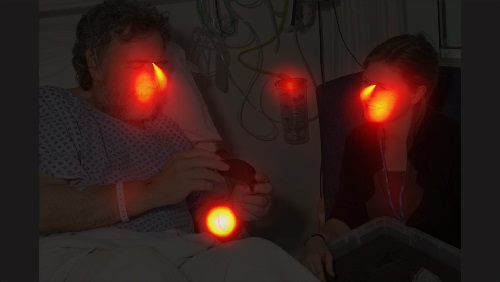}}
    \hfill
    \subfloat{
    \includegraphics[width=0.15\textwidth]{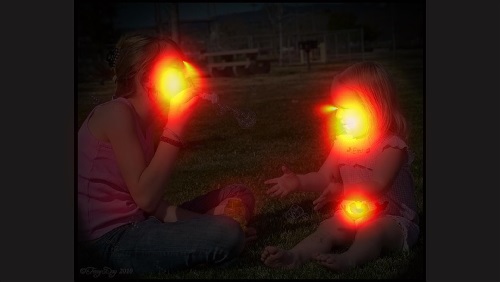}}
    \hfill
    \subfloat{
    \includegraphics[width=0.15\textwidth]{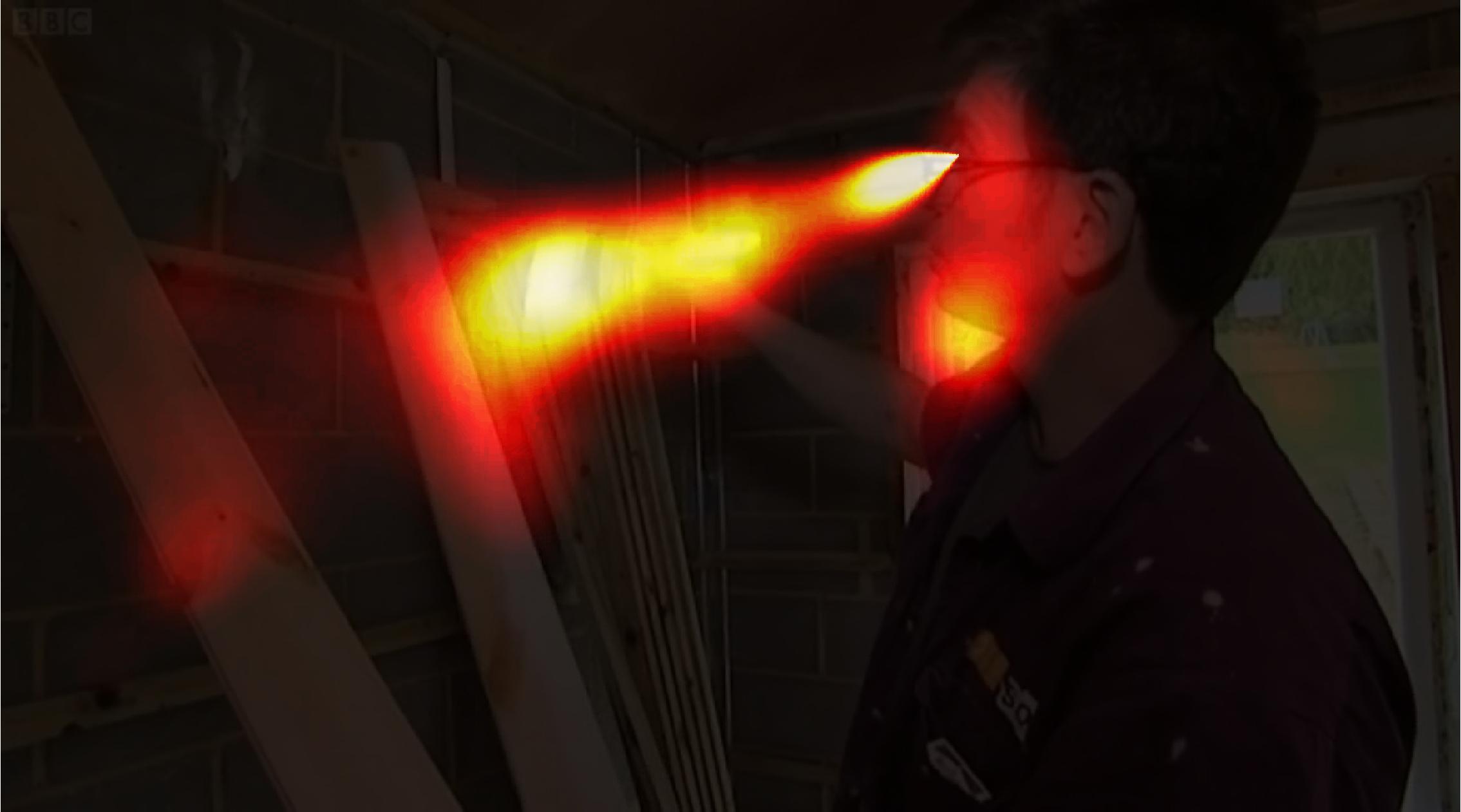}}
    \hfill
    \subfloat{
    \includegraphics[width=0.15\textwidth]{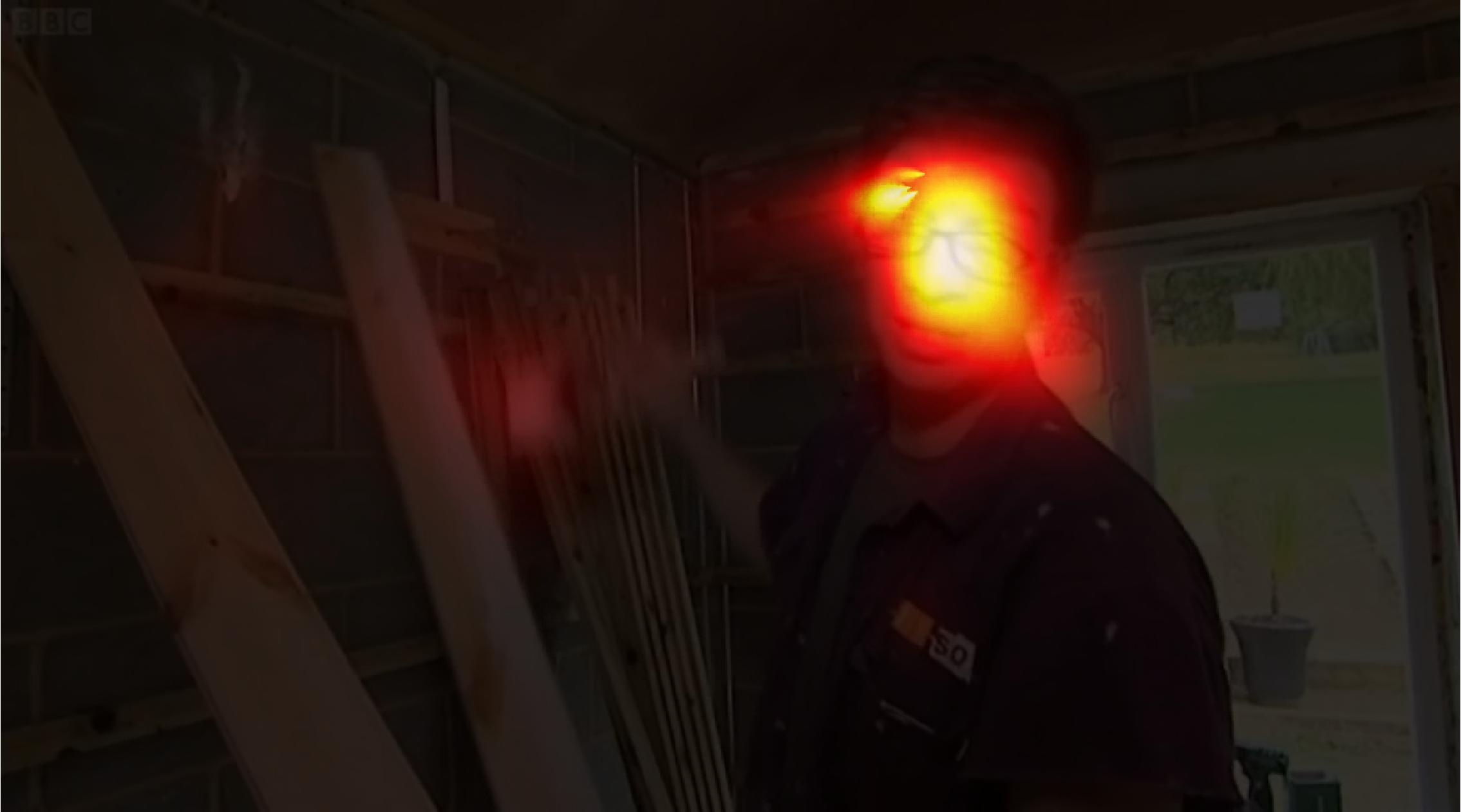}}
    \hfill
    \subfloat{
    \includegraphics[width=0.15\textwidth]{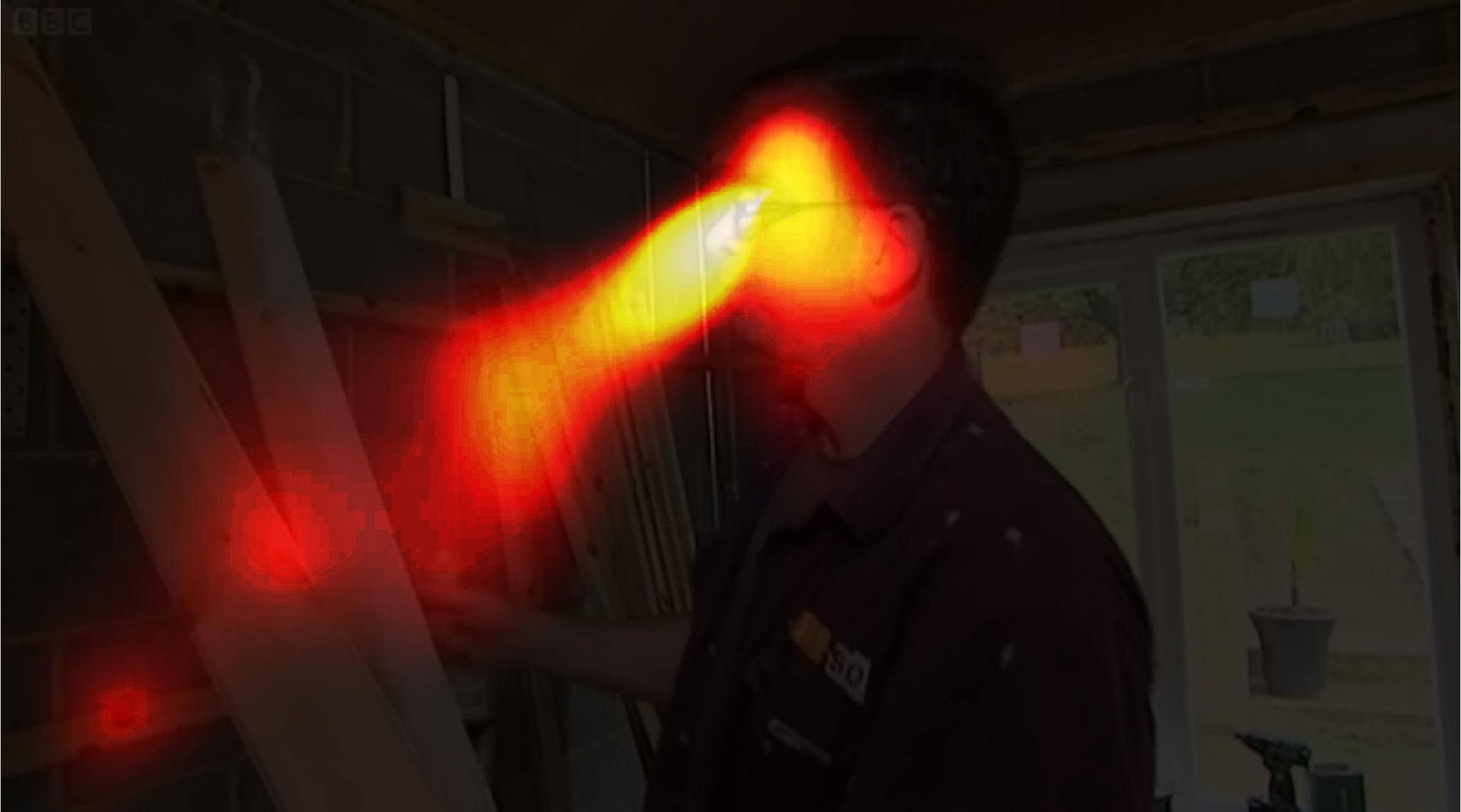}}
	\hfill\null	
\\[-1.7\baselineskip]	
    \centering
    \null\hfill
    {\tiny{(g)}}
    \hfill
    \subfloat{
    \includegraphics[width=0.15\textwidth]{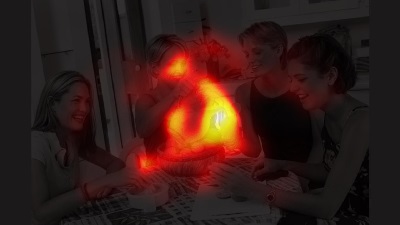}}    
    \hfill
    \subfloat{
    \includegraphics[width=0.15\textwidth]{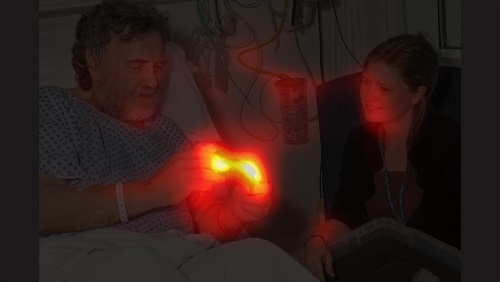}}
    \hfill
    \subfloat{
    \includegraphics[width=0.15\textwidth]{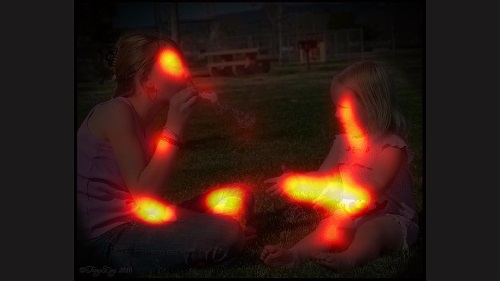}}
    \hfill
    \subfloat{
    \includegraphics[width=0.15\textwidth]{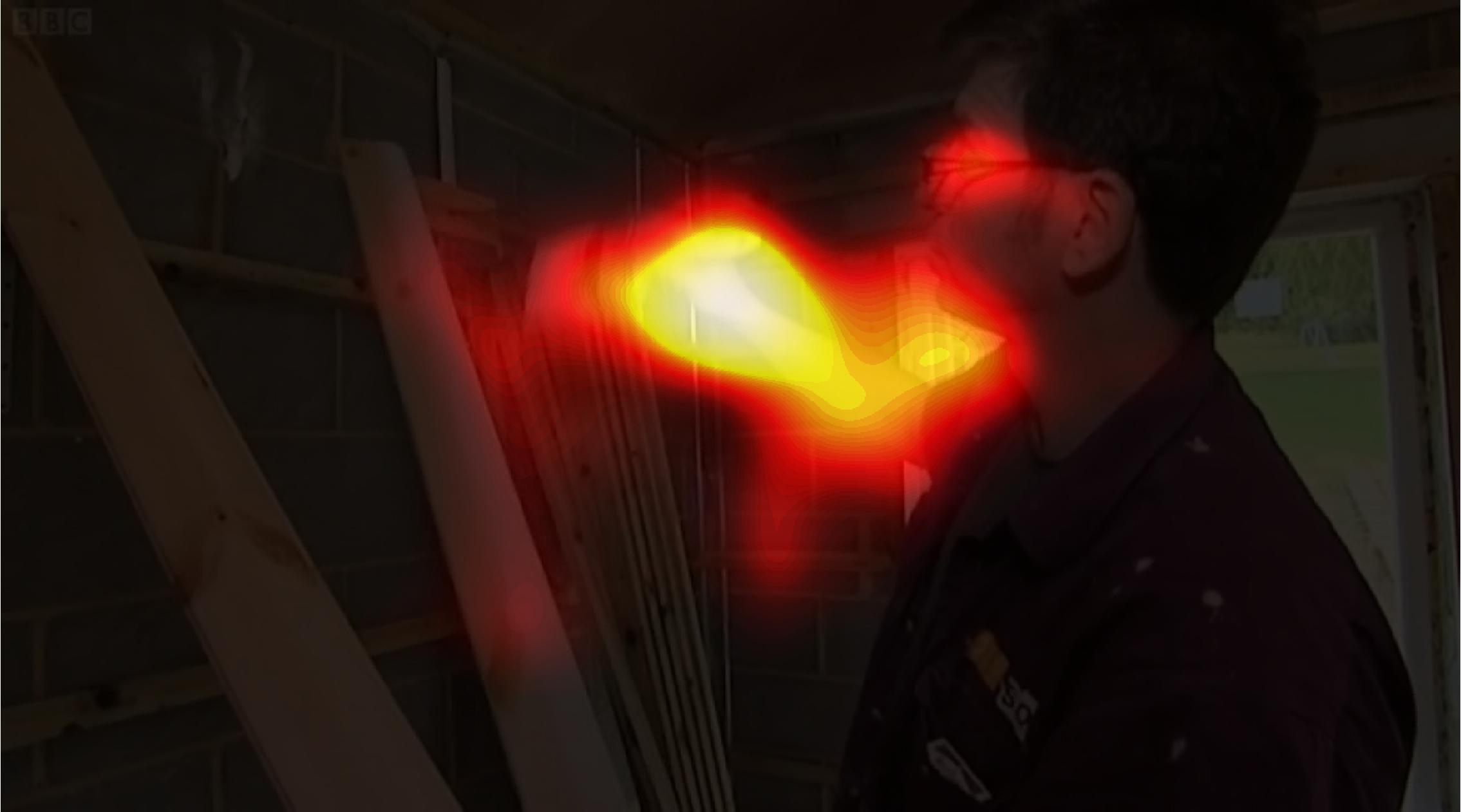}}
    \hfill
    \subfloat{
    \includegraphics[width=0.15\textwidth]{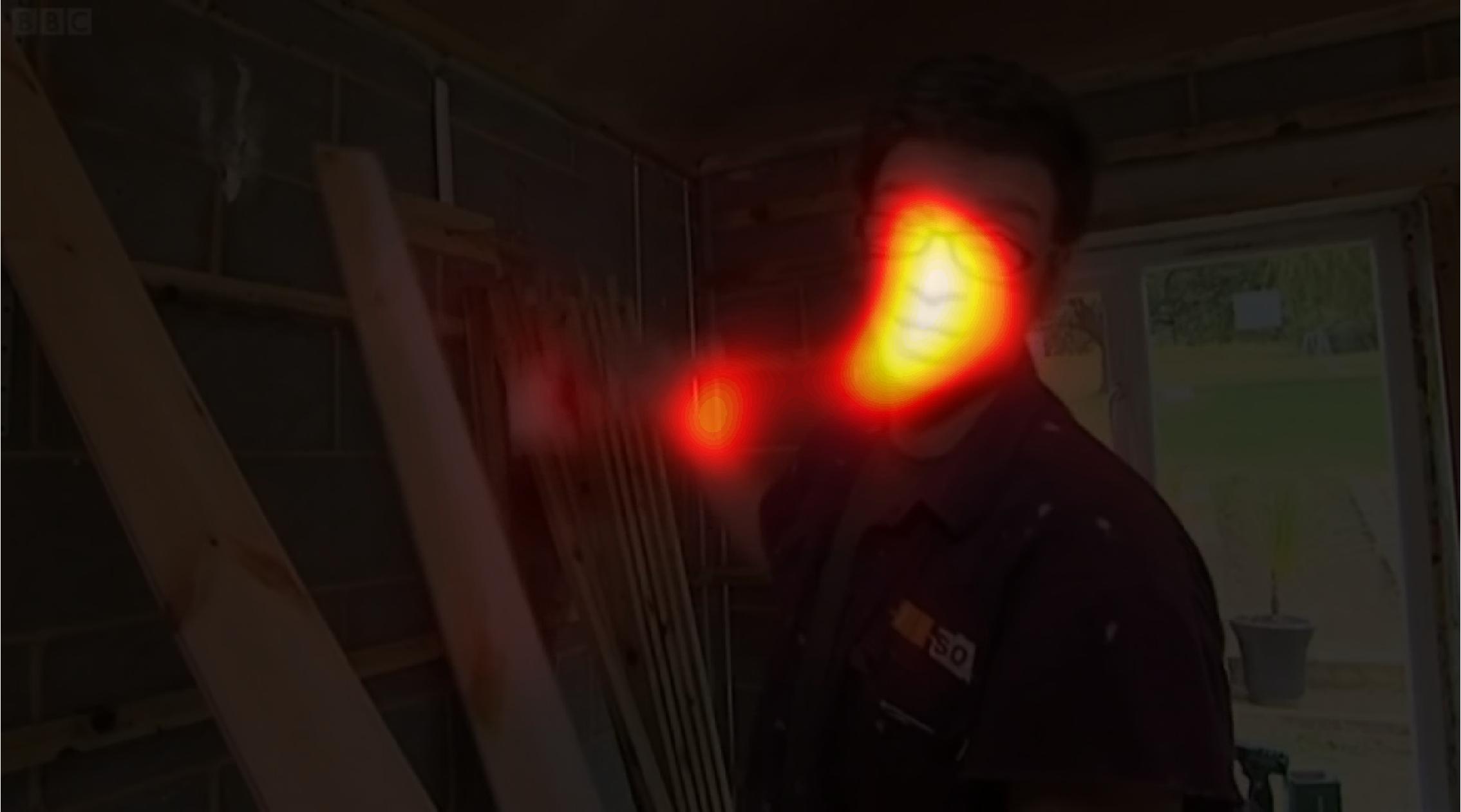}}
    \hfill
    \subfloat{
    \includegraphics[width=0.15\textwidth]{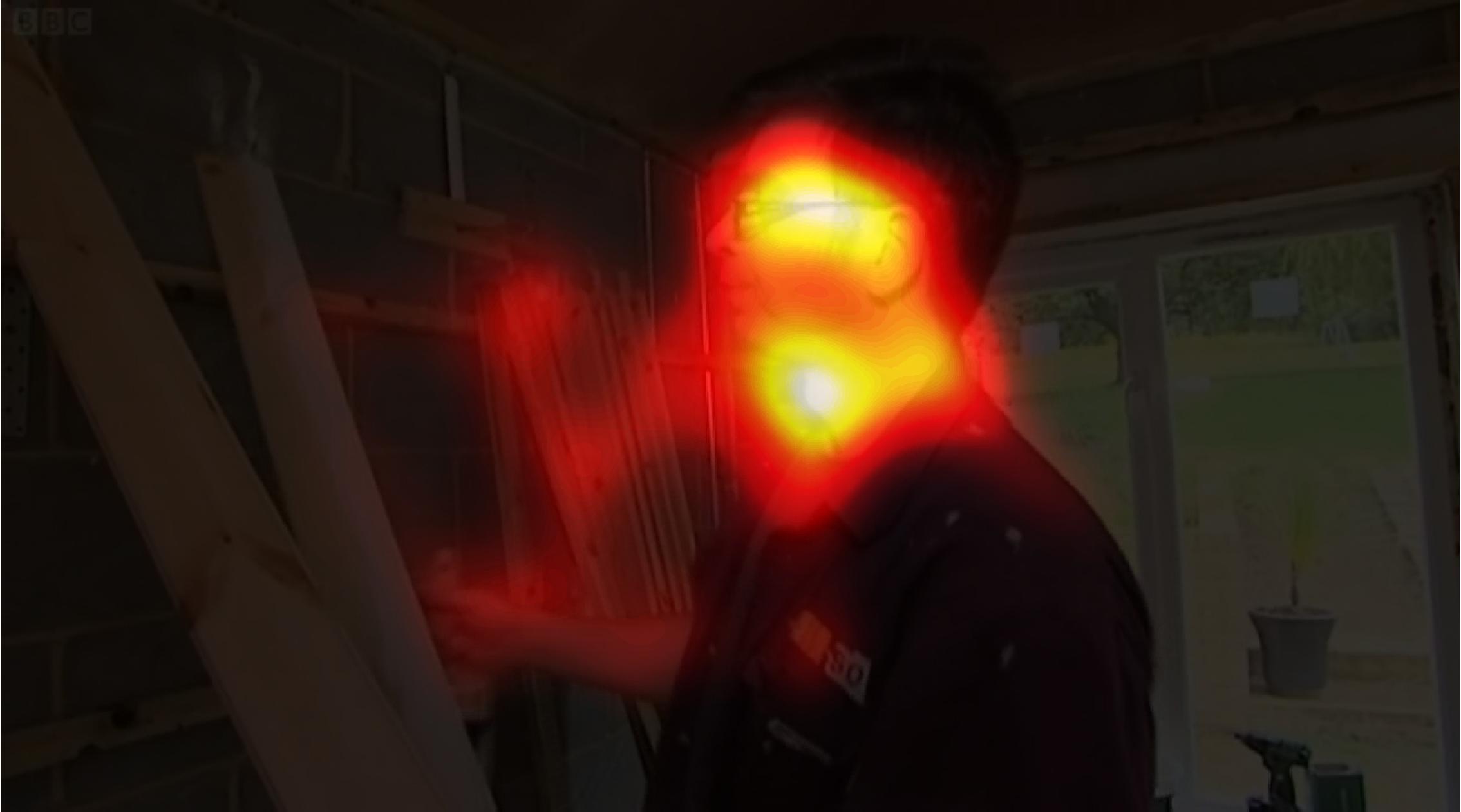}}
	\hfill\null
\\[-1.7\baselineskip]	
    \centering
    \null\hfill
    {\tiny{(h)}}
    \hfill
    \subfloat{
    \includegraphics[width=0.15\textwidth]{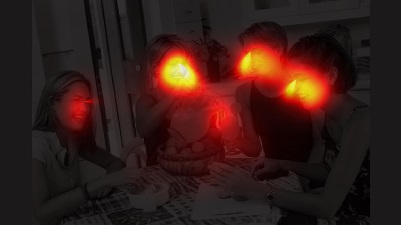}}    
    \hfill
    \subfloat{
    \includegraphics[width=0.15\textwidth]{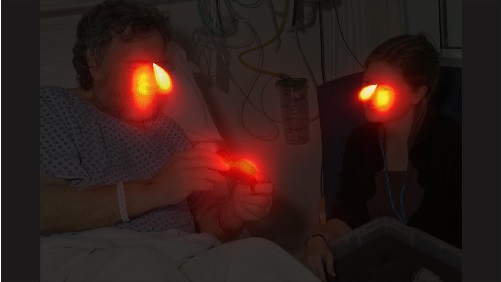}}
    \hfill
    \subfloat{
    \includegraphics[width=0.15\textwidth]{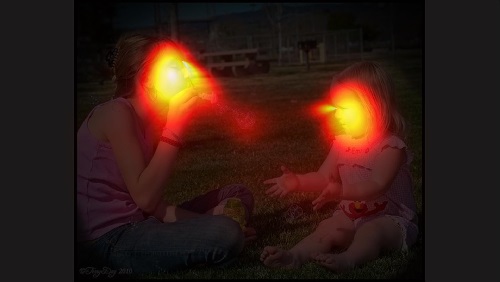}}
    \hfill
    \subfloat{
    \includegraphics[width=0.15\textwidth]{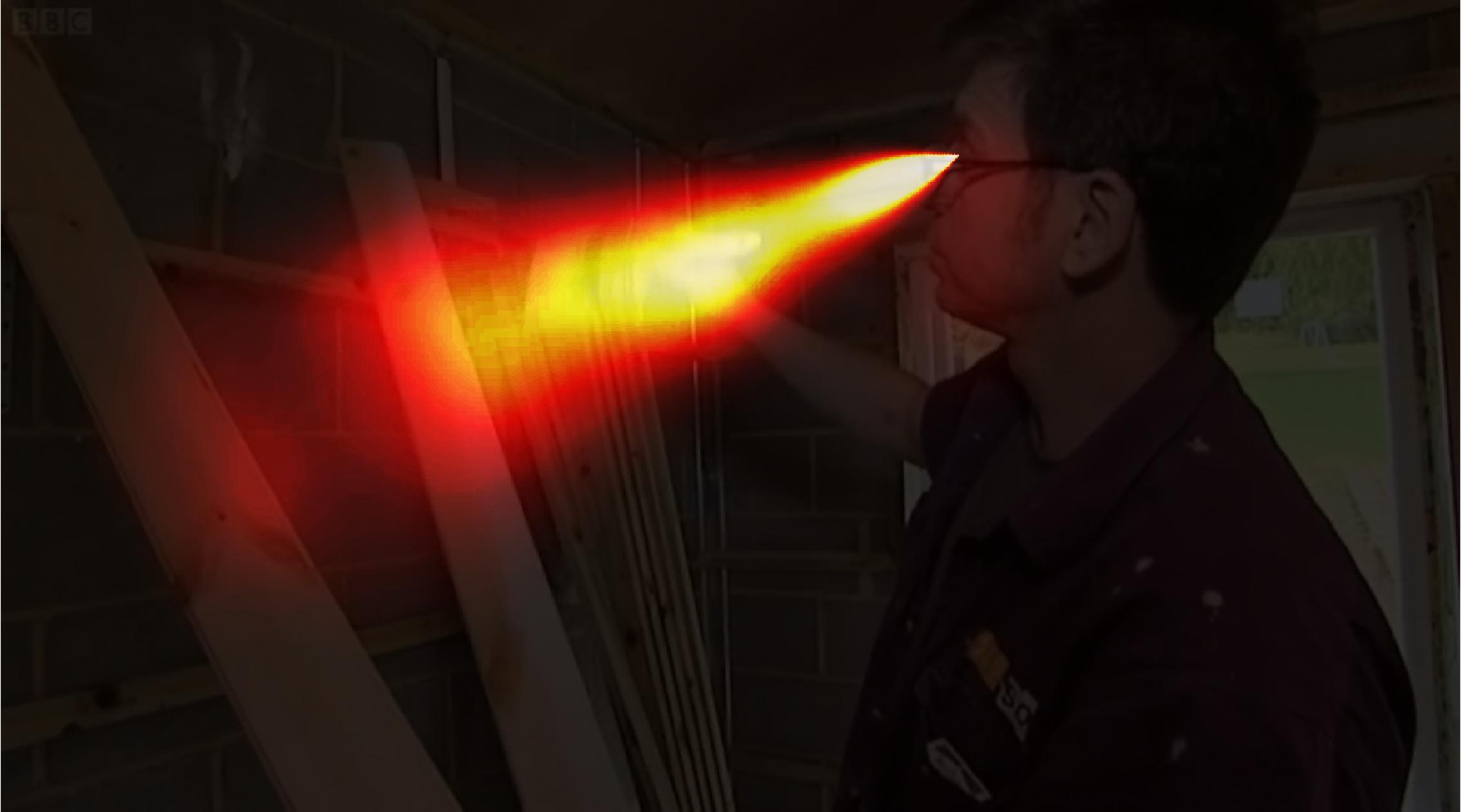}}
    \hfill
    \subfloat{
    \includegraphics[width=0.15\textwidth]{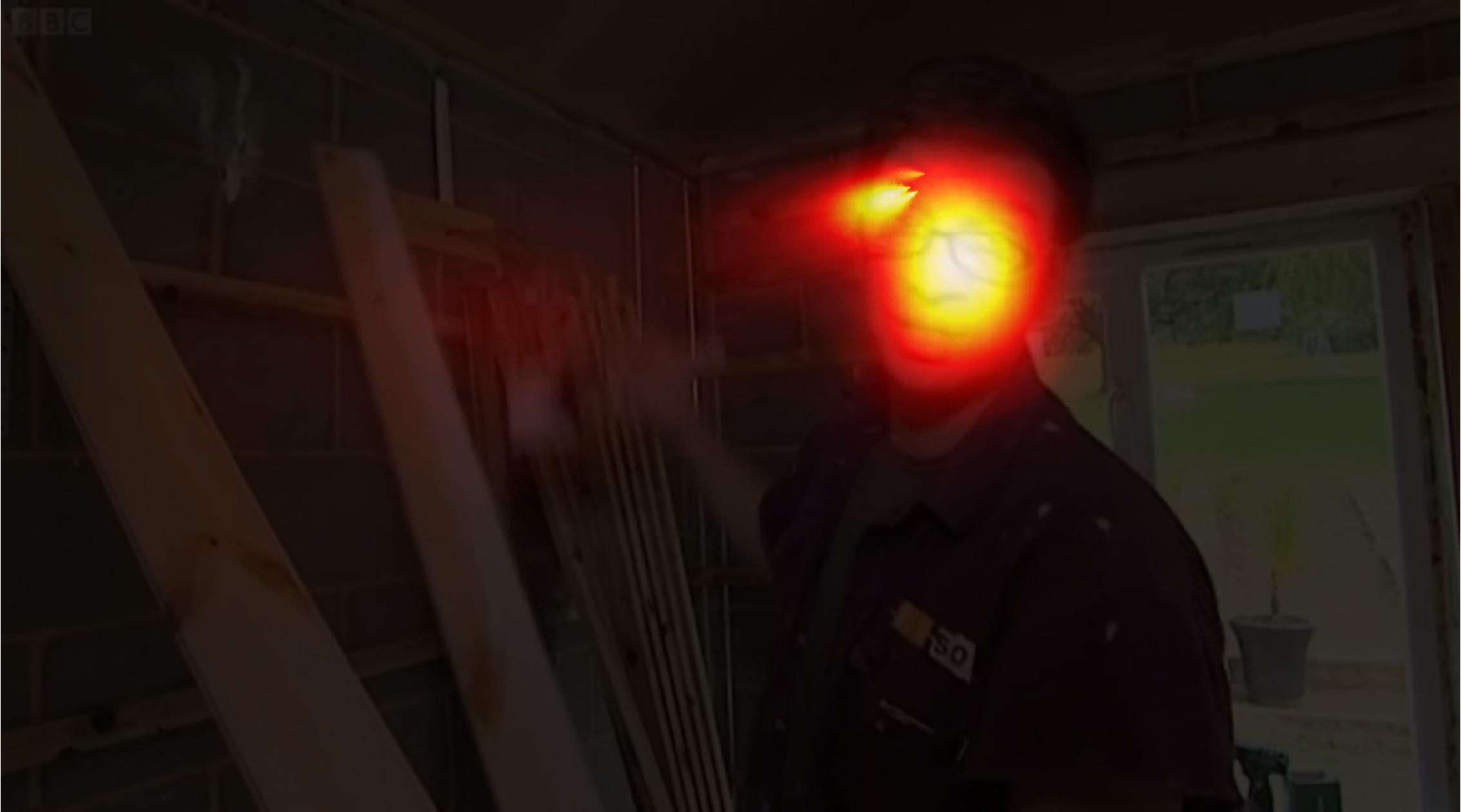}}
    \hfill
    \subfloat{
    \includegraphics[width=0.15\textwidth]{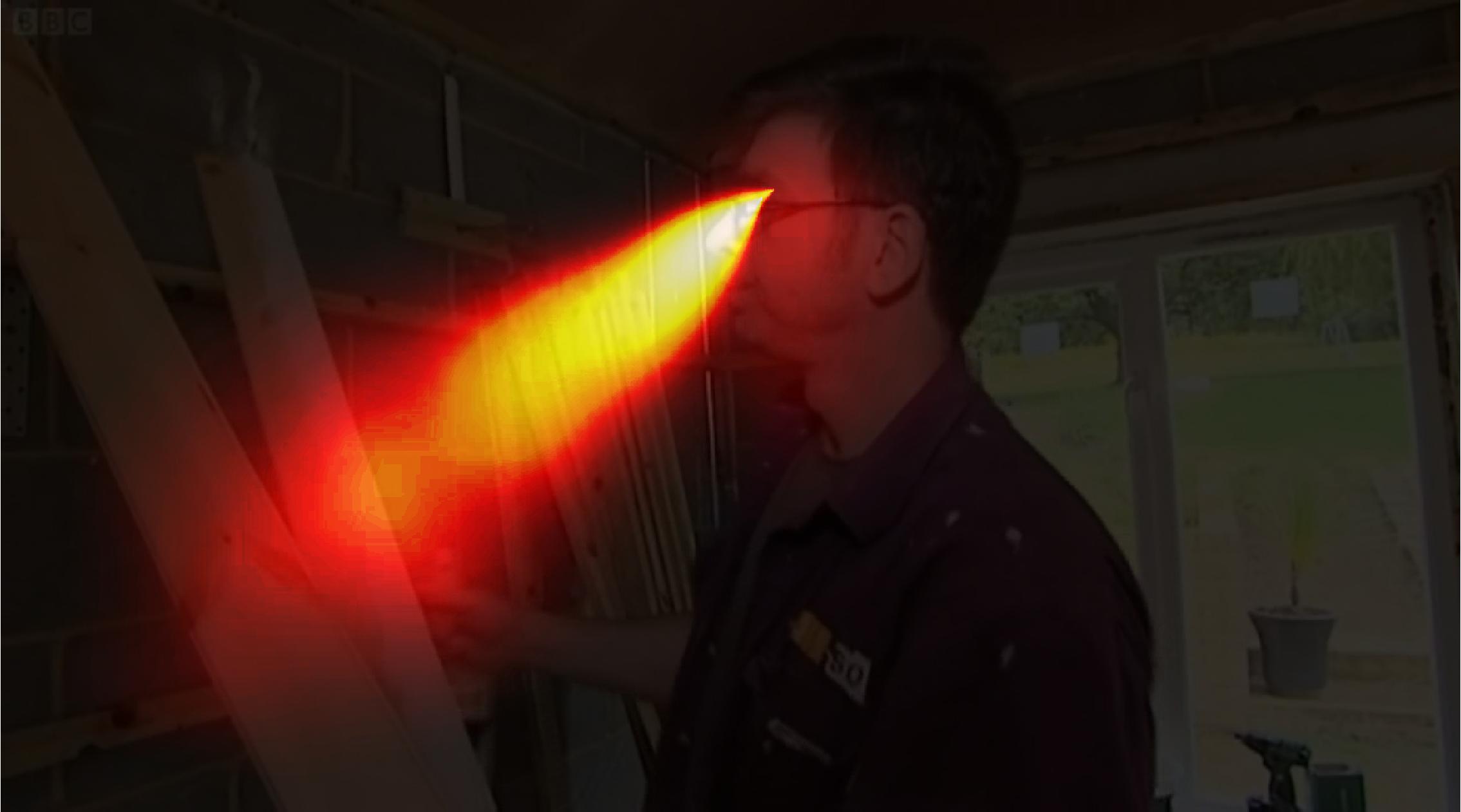}}
	\hfill\null	
\\[-1.7\baselineskip]	
    \centering
    \null\hfill
    {\tiny{(i)}}
    \hfill
    \subfloat{
    \includegraphics[width=0.15\textwidth]{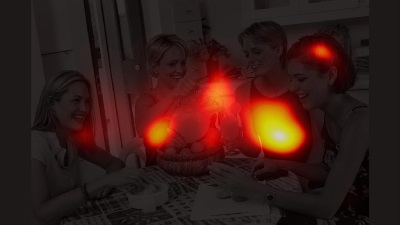}}    
    \hfill
    \subfloat{
    \includegraphics[width=0.15\textwidth]{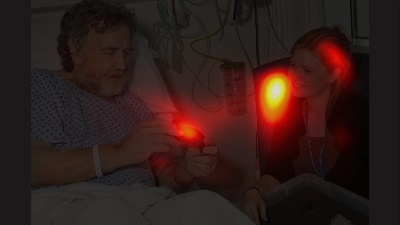}}
    \hfill
    \subfloat{
    \includegraphics[width=0.15\textwidth]{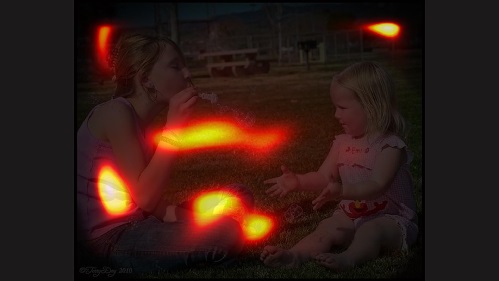}}
    \hfill
    \subfloat{
    \includegraphics[width=0.15\textwidth]{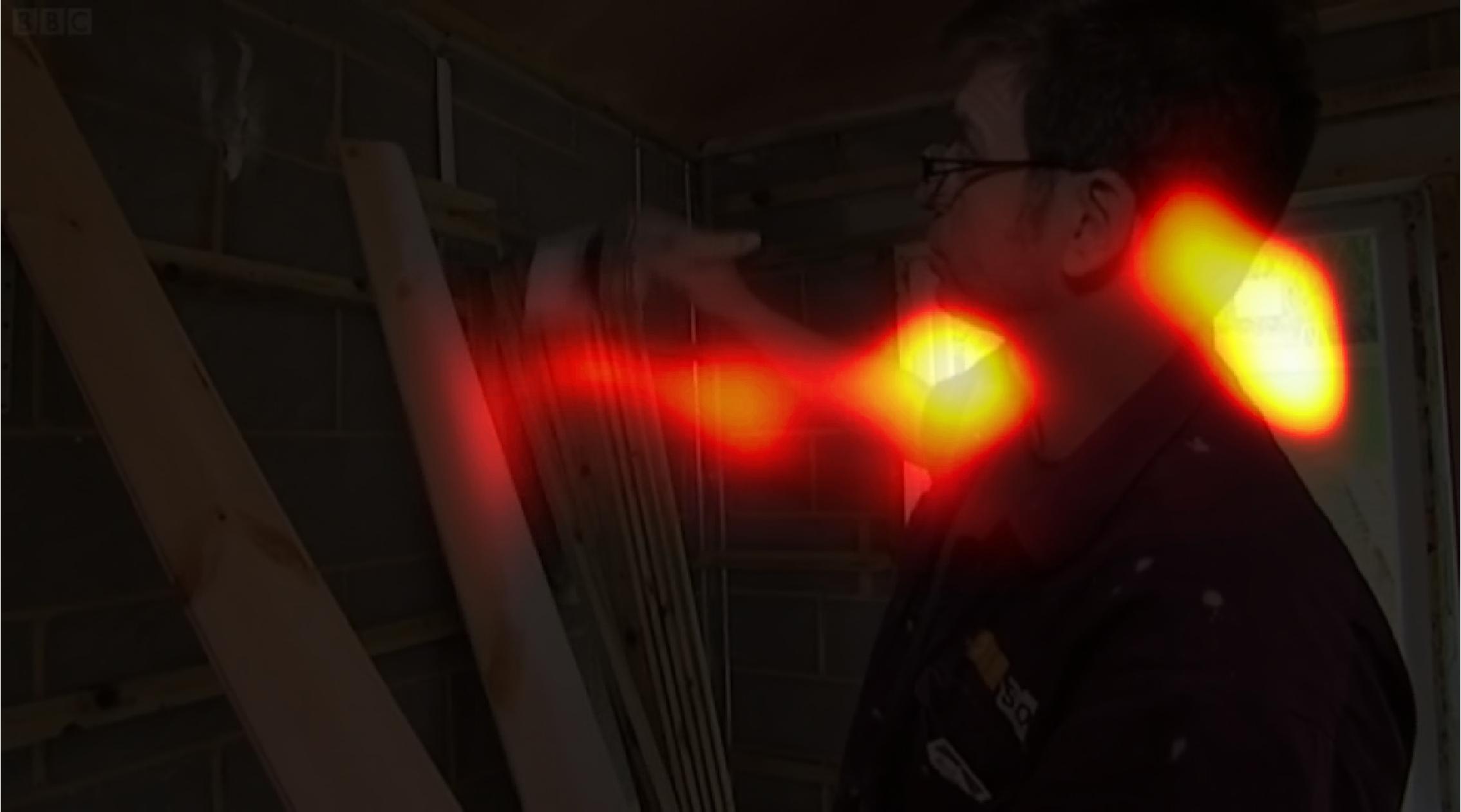}}
    \hfill
    \subfloat{
   \includegraphics[width=0.15\textwidth]{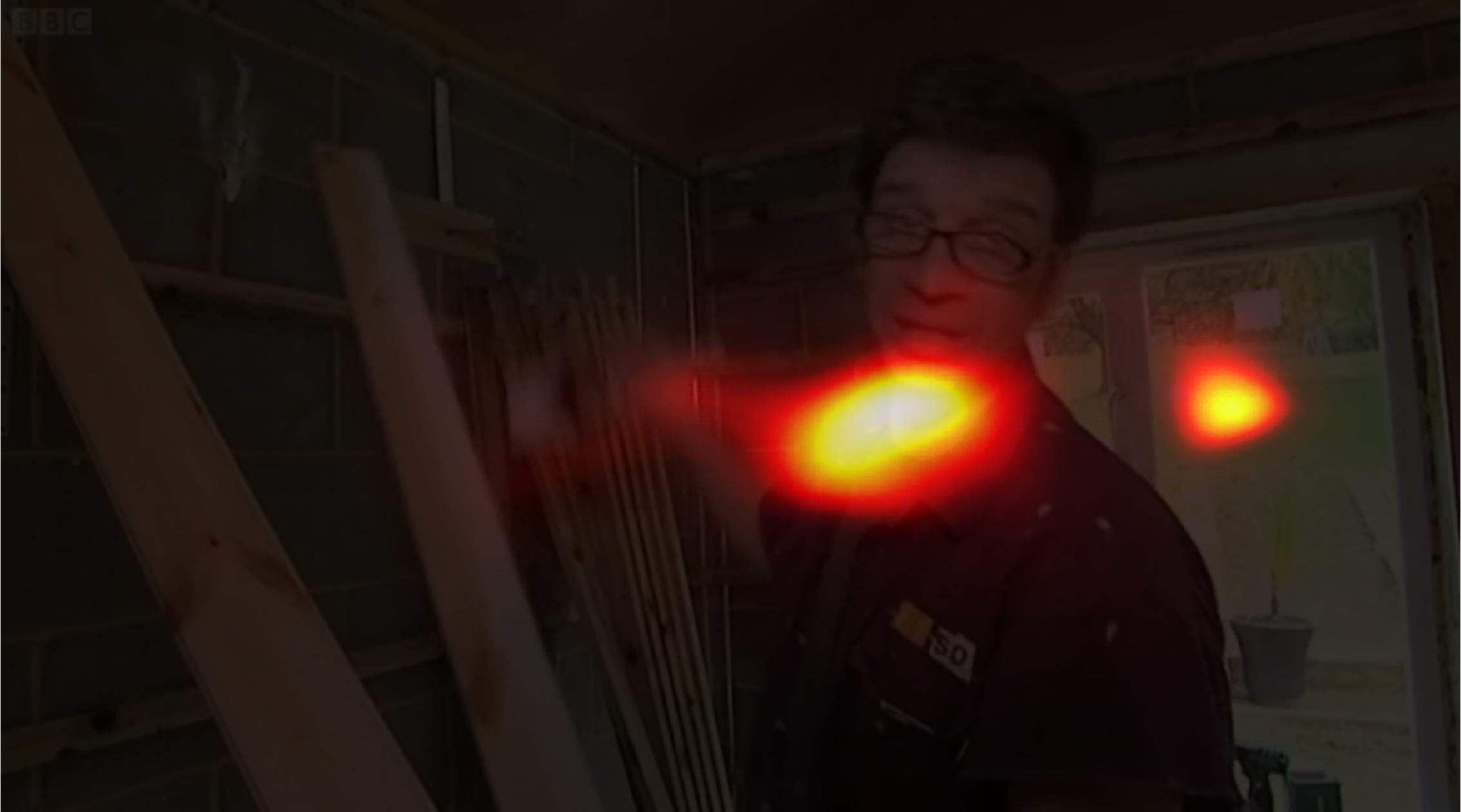}}
    \hfill
    \subfloat{
    \includegraphics[width=0.15\textwidth]{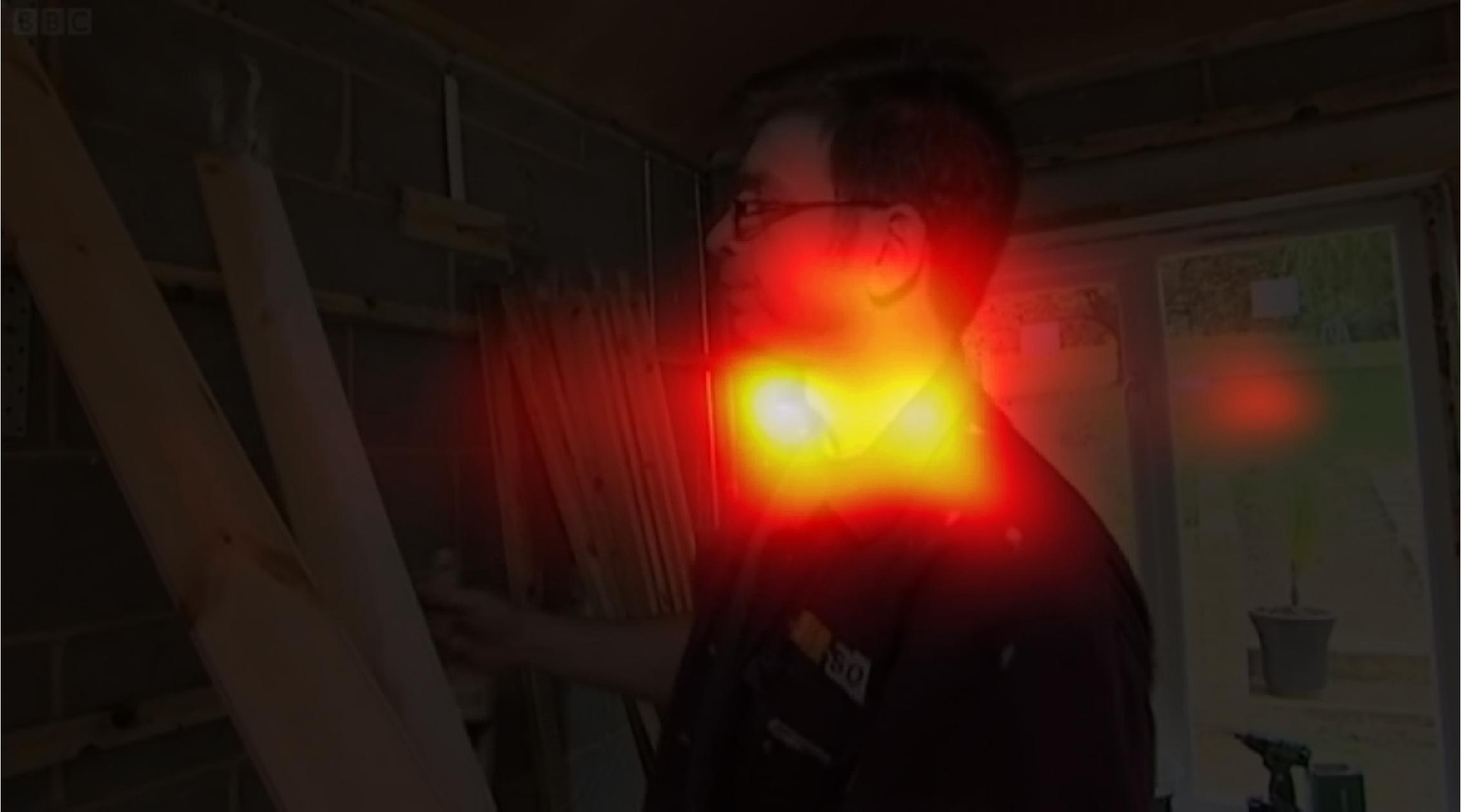}}
	\hfill\null
\\[-1.7\baselineskip]	
    \centering
    \null\hfill
    {\tiny{(j)}}
    \hfill
    \subfloat{
    \includegraphics[width=0.15\textwidth]{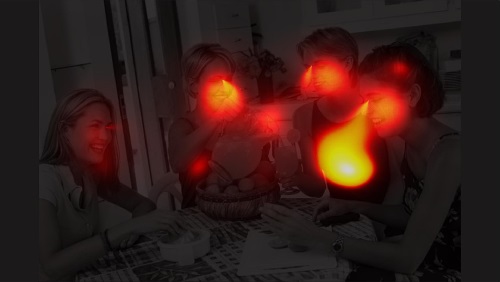}}    
    \hfill
    \subfloat{
    \includegraphics[width=0.15\textwidth]{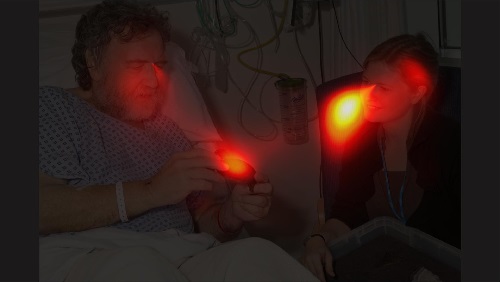}}
    \hfill
    \subfloat{
    \includegraphics[width=0.15\textwidth]{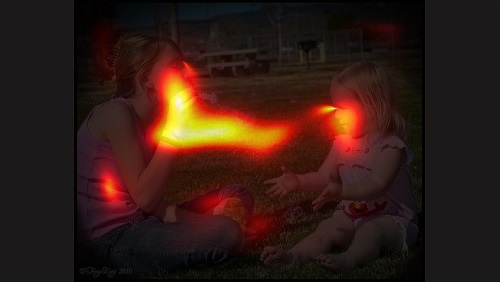}}
    \hfill
    \subfloat{
    \includegraphics[width=0.15\textwidth]{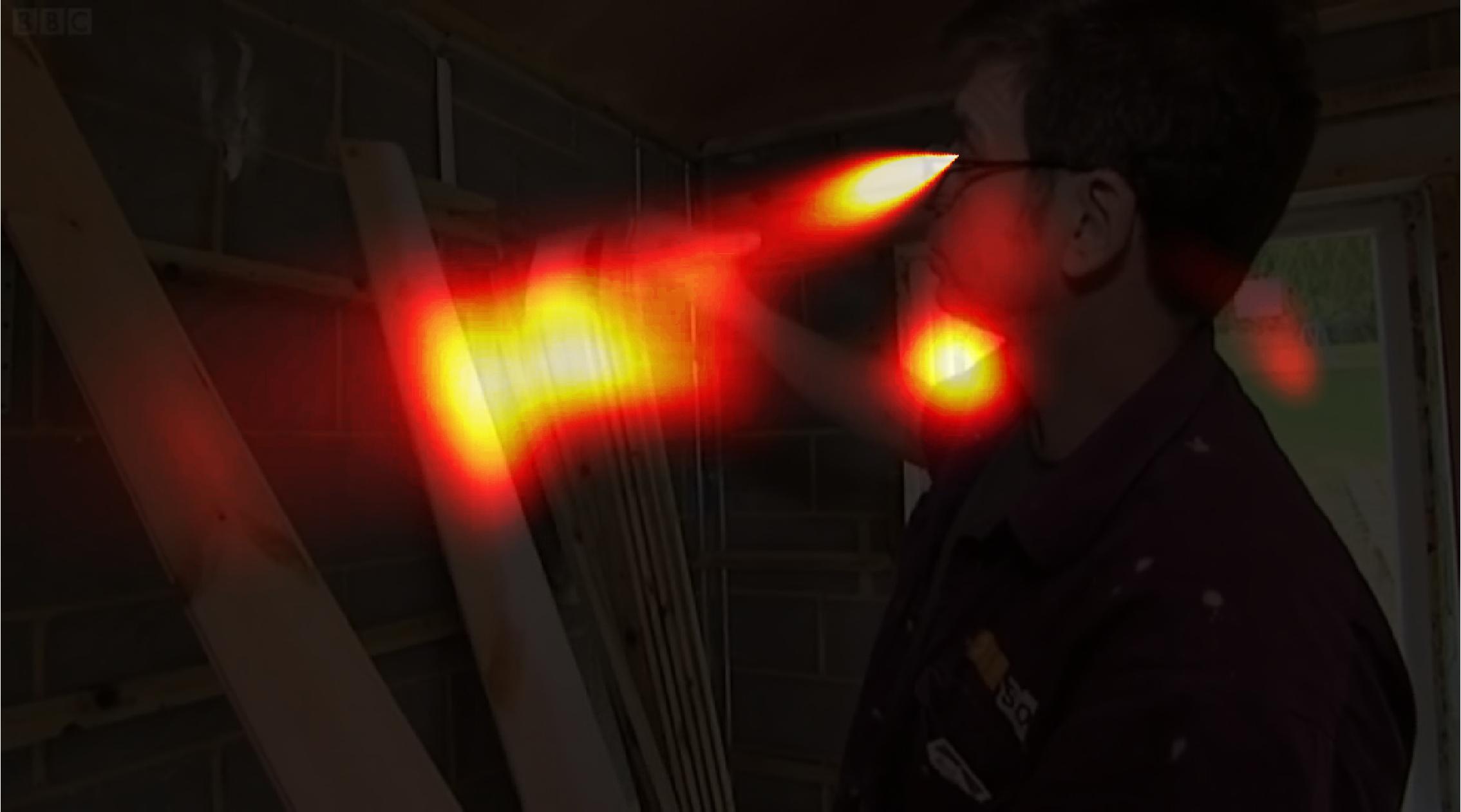}}
    \hfill
    \subfloat{
    \includegraphics[width=0.15\textwidth]{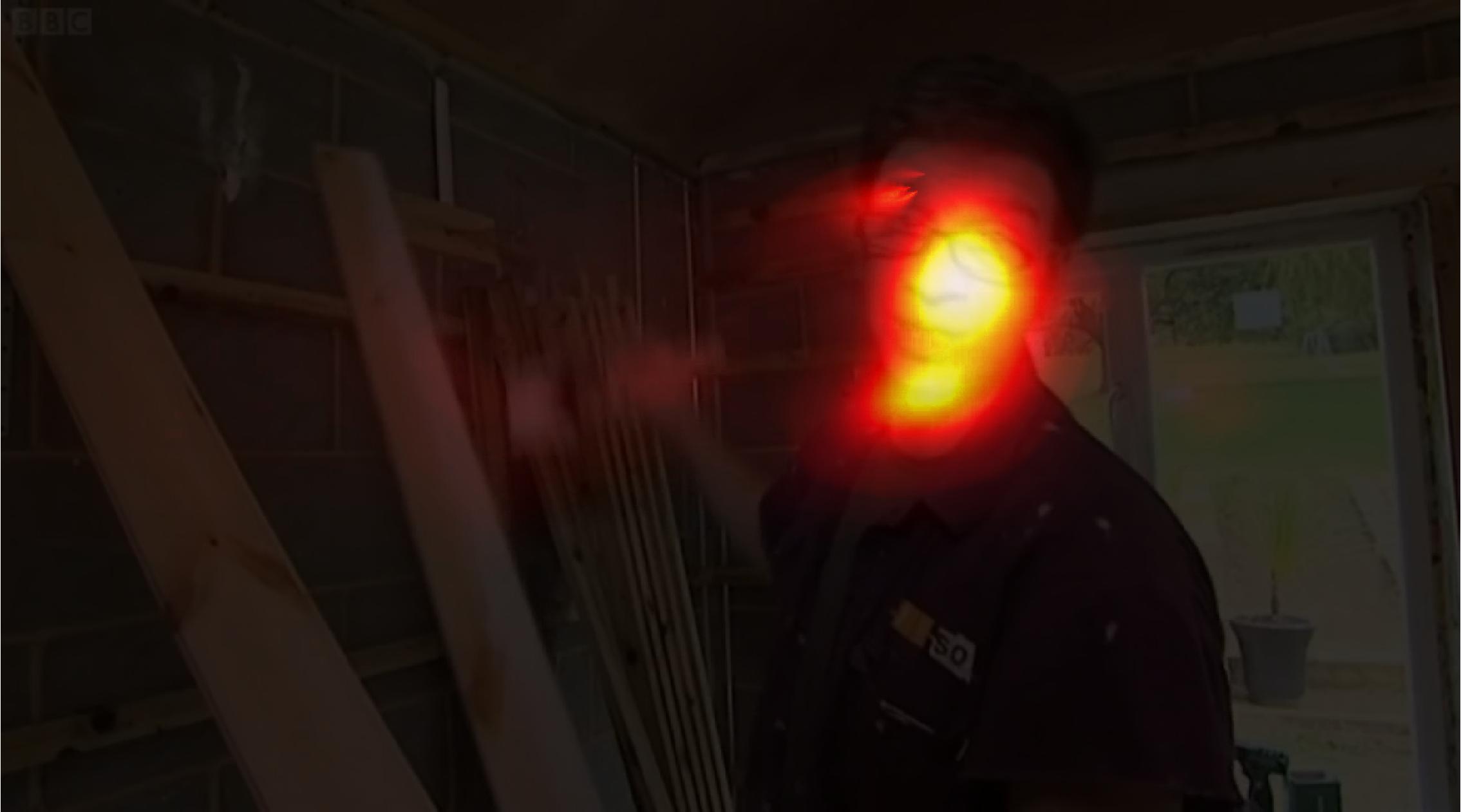}}
    \hfill
    \subfloat{
   \includegraphics[width=0.15\textwidth]{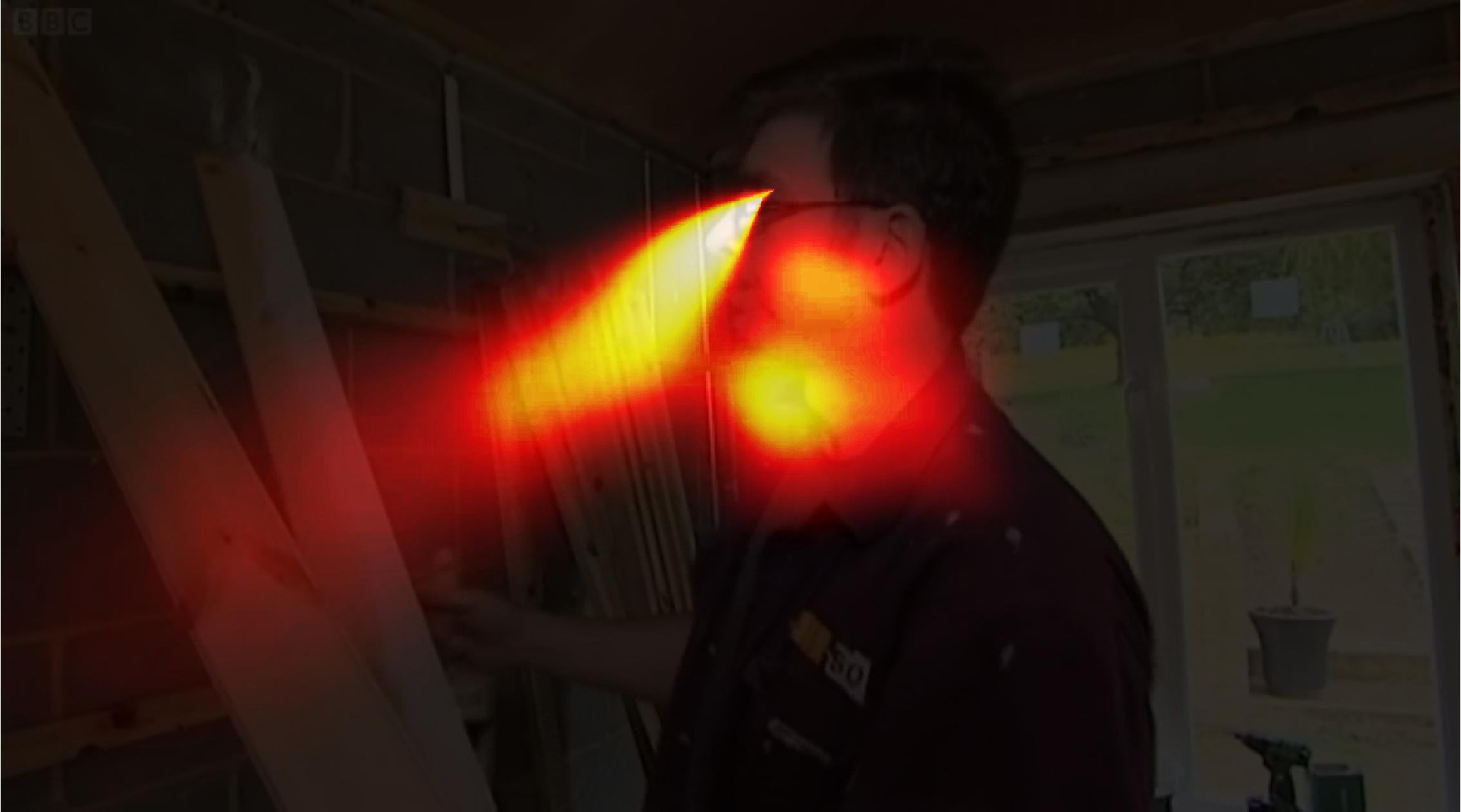}}
	\hfill\null							
	\setlength{\belowcaptionskip}{-10pt}
    \caption{\label{fig:final}
Sample images and video frames from the CAT2000 \cite{BorjiI15} and the DIEM \cite{DIEM} dataset with overlaid ground truth, overlaid saliency maps and overlaid Attentional Push-based augmented saliency maps. (a) Original images, (b) overlaid ground truth, (c) overlaid AWS maps, (d) overlaid augmented AWS maps, (e) overlaid BMS maps, (f) overlaid augmented BMS maps, (g) overlaid eDN maps, (h) overlaid augmented eDN maps,(i) overlaid FES maps, (j) overlaid augmented FES maps. Augmented saliency methods alter the standard saliency maps to be more consistent with the ground truth.}
\end{figure}

The performance improvements are more noticeable for the dynamic imagery. The average performance boosts for all of the augmented methods are 0.10, 1.19 and 0.18 for AUC, NSS and CC, respectively. The most significant performance boost in AUC score for the dynamic stimuli belongs to the augmented ContextAware model, which is more than 3 times larger than its improvement for static stimuli. 
This implies that the Attentional Push cues have more influence upon the viewers' fixation in dynamic scenes, which could be explained by the observation that the viewers feel more immersed while watching dynamic scenes. Example saliency maps for some of the augmented and standard saliency methods are shown in Fig. \ref{fig:final}.

\begin{table}
\centering
\caption{Average evaluation scores of  five separate augmented saliency maps, each based on a single Attentional Push cue for the dynamic stimuli.}
\label{table:details}
\begin{tabular}{lccccccccccccc}
\toprule    
• & None & \multicolumn{2}{c}{Static cues} &  \multicolumn{3}{c}{Dynamic cues} & All\\
    & & head/body pose & centerbias &  head/body pose & Bounce & SceneChange &  \\
    \midrule
    AUC   	& 0.79 & 0.87 & 0.82 & 0.80 & 0.80 & 0.81 & 0.91 \\
    NSS   	& 1.02 & 1.53 & 1.32 & 1.28 & 1.19 & 1.13 & 2.44 \\
	CC    	& 0.16 & 0.23 & 0.19 & 0.20 & 0.17 & 0.17 & 0.37 \\         
\bottomrule
\end{tabular}
\end{table}

To evaluate the effect of each Attentional Push cue in predicting the viewers' fixation, we create five separate augmented saliency maps, each based on a single Attentional Push cue. We use the AWS model as the standard saliency method to compute the augmented saliency maps. Table \ref{table:details} presents the average evaluation scores for the dynamic stimuli. Although the static Attentional Push cues seem to dominate most of the performance improvements, the dynamic Attentional Push cues have contribution in the performance improvements nonetheless. It should be noted that dynamic Attentional Push cues are not active in each frame and they require triggering event such as scene changes and changes in gaze direction. Given a saliency method augmented using only a dynamic Attentional Push cue, we can expect the average improvements over all the video frames to be small. Nevertheless, for a  saliency map augmented using a combination of static and dynamic Attentional Push cues, the dynamic cues can make contributions in improving the performance on many video frames that would be missed by static cues. It can be seen in Table \ref{table:details} that the combination of static and dynamic cues clearly outperforms static cues. 

We examined the cases in which the prediction performance of the augmented saliency map is lower than the saliency map in static stimuli. For each static stimulus, we consider images for which at least two of the three evaluation scores display degraded performance. There are twelve such images in total, with two of them showing degraded performance consistently in all evaluation metrics.
Both of these images contain crowded scenes, in which the actors are looking in many different directions. The reason for the degraded performance lies in the fact that the scene actors do not share the same loci of attention and therefore, the Attentional Push cues arising from their gaze directions compete with one another in pushing the viewers' attention. This situation leads to an inconsistent increase in the saliency values of many image regions that are not foci of actors' attention, which would lead to a degraded prediction performance for the augmented saliency method.

\section{Conclusion}
\label{Conclusion}
We presented an attention modeling scheme which combines Attentional Push cues, i.e. the power of image regions to direct and manipulate the attention allocation of the viewer, with standard saliency models, which generally concentrate on analyzing image regions for their power to pull attention. 
Our methodology significantly outperforms saliency methods in predicting the viewers' fixations on both static and dynamic stimuli. Our results showed that by employing Attentional Push cues, the augmented saliency maps can challenge the state of the art in saliency models.

\bibliographystyle{splncs}
\bibliography{eccv2016submission}
\end{document}